%% file: acl_latex.tex
\pdfoutput=1

\documentclass[11pt]{article}

\usepackage[final]{acl}

\usepackage{times}
\usepackage{latexsym}

\usepackage[T1]{fontenc}

\usepackage[utf8]{inputenc}

\usepackage{microtype}

\usepackage{inconsolata}

\usepackage{graphicx}
\usepackage{booktabs}
\usepackage{multirow}
\usepackage{multicol}
\usepackage{enumitem}
\usepackage{amsmath}
\usepackage{microtype}
\usepackage{caption}
\usepackage{wrapfig}
\usepackage{algorithm}
\usepackage{algpseudocode}
\usepackage{listings}
\usepackage{xcolor,colortbl}
\usepackage{csquotes}
\usepackage{pifont}
\usepackage{arydshln}
\usepackage{framed}
\usepackage{subcaption}
\usepackage{pgfplots}
\usepackage{pgfplotstable}
\usepackage{makecell}
\pgfplotsset{compat=1.18}
\usepgfplotslibrary{polar}
\usetikzlibrary{backgrounds,patterns,shapes}
\title{TCP: a Benchmark for Temporal Constraint-Based Planning}

\author{Zifeng Ding\thanks{Equal contribution.}$^1$, Sikuan Yan\footnotemark[1]$^2$, Zhangdie Yuan\footnotemark[1]$^1$,\\ \textbf{Xianglong Hu$^3$, Fangru Lin$^4$, Andreas Vlachos\thanks{Corresponding author.}$^1$} \\
$^{1}$University of Cambridge
$^{2}$LMU Munich \\ $^{3}$Amazon $^{4}$University of Oxford\\
\texttt{\{zd320, av308\}@cam.ac.uk}\\
}

\begin{document}
\maketitle
\begin{abstract}
Temporal reasoning and planning are essential capabilities for large language models (LLMs), yet most existing benchmarks evaluate them in isolation and under limited forms of complexity. To address this gap, we introduce the Temporal Constraint-based Planning (TCP) benchmark, that jointly assesses both capabilities. 
Each instance in TCP features a naturalistic dialogue around a collaborative project, where diverse and interdependent temporal constraints are explicitly or implicitly expressed, and models must infer an optimal schedule that satisfies all constraints. To construct TCP, we generate abstract problem prototypes 
that are then
paired with realistic scenarios from various domains and enriched into dialogues using an LLM. A human quality check is performed on a sampled subset to confirm the reliability of our benchmark. We evaluate state-of-the-art LLMs and find that even the strongest models may struggle with TCP, highlighting its difficulty and revealing limitations in LLMs’ temporal constraint-based planning abilities. We analyze underlying failure cases, open source our benchmark\footnote{https://huggingface.co/datasets/Beanbagdzf/TCP}, and hope our findings can inspire future research. 
\end{abstract}

\section{Introduction}
With the rise of autoregressive large language models (LLMs), recent studies have aimed to evaluate their temporal reasoning abilities. \citet{DBLP:conf/naacl/Qiu0ZKPC24} initially found that LLMs lack temporal grounding and struggle with even simple reasoning tasks. Subsequent benchmarks 
have aimed to assess temporal reasoning more broadly, and newer models have demonstrated improved performance on basic skills such as calculations and comparisons involving time, suggesting that temporal grounding has improved over model generations \cite{DBLP:conf/acl/Wang024}. 
Despite this research, it remains unclear whether LLMs can handle more complex forms of temporal reasoning, such as understanding and reasoning under temporal constraints.
Existing benchmarks often fall short in this regard, and two key limitations remain. (1) They primarily focus on simple tasks and do not assess reasoning over multiple interacting temporal constraints \citep{DBLP:conf/acl/ChuCCY00024,DBLP:conf/acl/XiongPKF24,DBLP:conf/acl/Wang024,DBLP:conf/iclr/FatemiKTMYPSHP25}. 
(2) Many benchmarks rely on questions from knowledge bases \citep{DBLP:conf/nips/ChenWWW21,DBLP:conf/acl/TanNB23,DBLP:journals/corr/abs-2501-03040}, which allows models to rely on memorized pre-training data rather than demonstrate genuine temporal reasoning \citep{DBLP:conf/iclr/FatemiKTMYPSHP25}.
\begin{figure*}
    \centering
    \includegraphics[width=\textwidth]{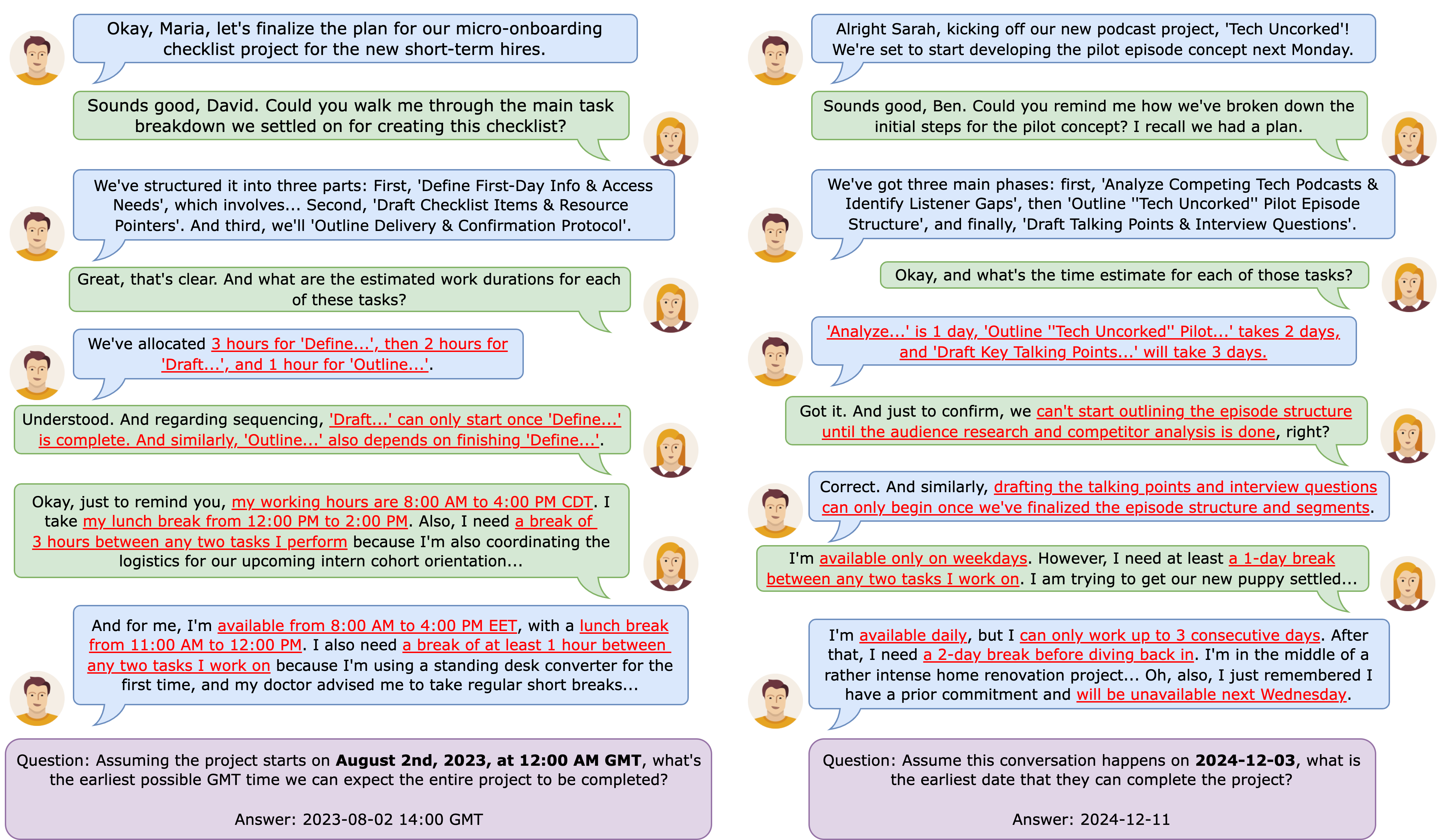}
    \caption{Example \textit{short} (\textbf{left}) and \textit{long} (\textbf{right}) problems in TCP. Some parts are abbreviated due to limited space. The red underlined contents correspond to the temporal constraints. The problem on the left belongs to the domain of Business Administration and involves the dependency type A $\rightarrow$ B, A $\rightarrow$ C. The problem on the right falls within the domain of Communications/Media Studies and involves the dependency type A $\rightarrow$ B, B $\rightarrow$ C. See Sec. \ref{sec: benchmark construction} for the details of involved domains and dependency types in TCP. See Appendix \ref{app: examples} for more example problems.}
    \label{fig:data_example}
    \vspace{-10pt}
\end{figure*}

In parallel, recent studies have started to examine LLMs’ ability to plan under various constraints, with a growing interest in incorporating temporal aspects. Some works explore travel planning \citep{DBLP:conf/icml/Xie0CZLTX024}, where time-related constraints are present, but the focus remains on producing plausible itineraries rather than testing deeper temporal reasoning. Others investigate calendar or meeting scheduling \citep{DBLP:journals/corr/abs-2406-04520}, often limited to straightforward calculations involving time. Further, Asynchow \citep{DBLP:conf/icml/LinMHY0P24} introduces overlapping task execution to evaluate how LLMs manage asynchronous planning, yet it primarily involves simple time addition. While these efforts highlight the relevance of temporal reasoning in planning, none offers a comprehensive benchmark specifically for evaluating LLMs' planning ability under diverse and interacting temporal constraints.

To this end, we introduce a new benchmark focused on \textbf{T}emporal \textbf{C}onstraint-based \textbf{P}lanning (\textbf{TCP}). Unlike prior benchmarks, TCP requires LLMs to generate plans under diverse and interdependent temporal constraints. Each instance is framed as a planning problem over a cooperative project, presented as a dialogue in which participants discuss scheduling under temporal constraints, followed by a question about the optimal plan.
To evaluate a range of temporal reasoning skills, we design two categories of problems: \textit{short} and \textit{long}, where the associated projects must be completed within 1 day and 1 week, respectively. See Fig.~\ref{fig:data_example} for examples from both categories.

To construct TCP, 
we adopt a scalable construction approach that first generates abstract prototypes and then enriches them using LLMs. We first generate problem prototypes using a Python script that samples temporal constraints (constraint types listed in 
Table \ref{tab:temporal_constraints}) from a predefined pool based on manually written templates. The script performs an exhaustive search to ensure logical correctness, and generates the answer for each instance. To enrich these abstract prototypes with realistic context and language, we manually author 30 domain-specific project scenarios across 10 real-world domains. These scenarios are then expanded using an LLM, 
with human experts reviewing and correcting outputs that do not meet quality standards. The final benchmark instances are created by pairing each prototype with a matching scenario and prompting the LLM to produce a 
dialogue that embeds the temporal constraints in context. To minimize the influence of pre-training data memorization, instances are fully anonymized: all task names and contextual details are generated from scratch and do not refer to real-world entities or events.

We ensure benchmark quality through a two-stage verification process: a symbolic consistency check that validates constraint preservation via structured metadata, and a semantic check assisted by an LLM and verified by humans to confirm alignment between the prototype and the final instance. 
To further validate the quality of our generated data, we conduct a final round of human annotation on a randomly sampled subset of the benchmark. 
All answers provided by the benchmark are verified by annotators, who independently arrive at the same answers given only the dialogues and questions, suggesting that our construction pipeline reliably produces high-quality instances and provides a practical solution for scaling datasets on cooperative project planning.
\begin{table*}[ht]
\centering
\resizebox{0.85\textwidth}{!}{
\begin{tabular}{lp{11cm}}
\toprule
\textbf{Temporal Constraint \& Description} & \textbf{Example with Abbreviated Task Names in Dialogue} \\
\midrule
\texttt{Task Duration}: &  Ben: Yep, Task A should take about 2 days. Task B will be around \\ specify the duration of a task. &  3 days. And Task C is estimated at 2 days. \\
\midrule
\texttt{Task Dependency}: & Carter: It's crucial that Task B only starts after Task A is fully completed.\\ specify dependencies between tasks. & And Task C also needs Task A to be finished before we can proceed. \\ 
\midrule
\texttt{Break Between}: & David: Plus, I'll also need a minimum 2-day break between my tasks on \\ specify a participant's break between two performing tasks. & this project, as I'm managing the final integration testing for...\\
\midrule
\texttt{Break After}: & Eleanor: And I can manage up to 4 consecutive hours of work, but after \\ specify a participant's break after working for a duration. & that, I must take at least a 1-hour break to rest my eyes...\\
\midrule
\texttt{General Availability}: & Sarah: Just a reminder about my availability: I'm on AEST, working \\ specify the general available time. & 11:00 AM to 7:00 PM, with my lunch break from 3:00 PM to 4:00 PM.  \\
\midrule
\texttt{Specific Unavailability}: & Miller: Also, please remember I'm completely unavailable next Wed- \\ specify the specific unavailablity on top of general availability. & nesday – that's the day of the science fair, and I promised to be there... \\

\bottomrule
\end{tabular}
}
\caption{Temporal constraint types with abbreviated task names for brevity. Each problem category includes all constraints but varies in expression based on temporal scope. Examples are sampled from different categories.}
\label{tab:temporal_constraints}
\vspace{-10pt}
\end{table*}

To summarize, we have two key contributions: (1) We propose TCP, the first temporal constraint-based planning benchmark that specifically evaluates LLMs' ability in planning under interdependent temporal constraints. (2) We evaluate a range of state-of-the-art LLMs and find that even the most advanced reasoning models fall short of perfect accuracy, 
underscoring the difficulty of the benchmark and the remaining gap in LLMs' capabilities in temporal constraint-based planning. Our analysis reveals key failure modes, including difficulty with asynchronous scheduling and errors in time zone reasoning. We hope these findings can guide and inspire future research.
\section{Related Work}
\paragraph{Temporal Reasoning with LLMs.}
A number of benchmarks have been proposed to evaluate LLMs’ temporal reasoning abilities. \citet{DBLP:conf/naacl/Qiu0ZKPC24} first show that LLMs lack temporal grounding, often struggling even with simple temporal tasks. Building on this, subsequent work introduce benchmarks covering broader range of temporal reasoning skills \citep{DBLP:conf/acl/ChuCCY00024,DBLP:conf/acl/Wang024,DBLP:journals/corr/abs-2501-03040}. However, many benchmarks partly rely on real-world knowledge, allowing LLMs to answer by recalling memorized facts rather than reasoning. To address this, \citet{DBLP:conf/iclr/FatemiKTMYPSHP25} propose Test of Time, a synthetic benchmark designed to decouple temporal reasoning from factual knowledge. Similarly, TGQA \citep{DBLP:conf/acl/XiongPKF24} is constructed from knowledge graphs (KGs) with anonymized entities, thereby reducing the impact of memorized knowledge from pre-training. This also addresses a key limitation of earlier QA benchmarks over temporal KGs \citep{DBLP:conf/acl/SaxenaCT20,DBLP:conf/semweb/DingLQWHMMCLHT23}, where entities were not anonymized. While these efforts represent progress, they primarily focus on isolated or basic temporal skills and do not evaluate more complex forms of temporal reasoning, such as reasoning over multiple interacting temporal constraints as emphasized in our TCP benchmark.
\paragraph{Planning with LLMs.} 
Recent works have explored LLMs’ planning capabilities, beginning with PlanBench \citep{DBLP:conf/nips/ValmeekamMHSK23}, which focuses on goal-directed deterministic planning. Follow-up studies \citep{DBLP:journals/corr/abs-2302-05128,DBLP:conf/acl/YuanCFGSJXY23,DBLP:conf/icml/KambhampatiVGVS24,DBLP:conf/iclr/StechlyVK25} extend this line towards constraint-based planning. More recent efforts start to incorporate temporal constraints, such as travel planning \citep{DBLP:journals/corr/abs-2405-20625,DBLP:conf/icml/Xie0CZLTX024}, meeting scheduling in N\textsc{atural} P\textsc{lan} \citep{DBLP:journals/corr/abs-2406-04520}, and parallel task coordination \citep{DBLP:conf/icml/LinMHY0P24,DBLP:conf/acl/ZhangYH0XC24,DBLP:journals/corr/abs-2503-02238}. While these works show growing interest in temporal reasoning, they lack systematic evaluation of temporal constraint-based planning. Travel planning focuses on generating plausible itineraries rather than reasoning over temporal constraints; N\textsc{atural} P\textsc{lan} focuses on basic clock-time calculations; and works for parallel task coordination are often limited to time addition and task parallelism. In contrast, TCP requires more comprehensive temporal understanding, including time zones and dynamic unavailability (e.g., \texttt{Break Between} and \texttt{Break After} in Table \ref{tab:temporal_constraints}), making it a more targeted benchmark for evaluating LLMs’ temporal constraint-based planning abilities.
\begin{figure*}[t]
    \centering
    \includegraphics[width=\textwidth]{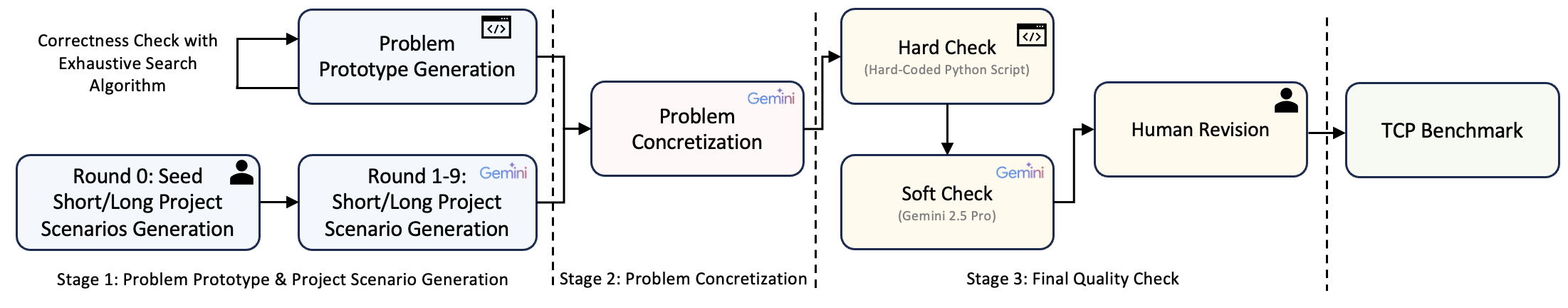}
    \caption{Overview of our benchmark generation pipeline.}
    \label{fig:construction}
    \vspace{-10pt}
\end{figure*}
\section{Temporal Constraint-Based Planning}
\subsection{Problem Design}
We aim to evaluate LLMs’ ability in temporal constraint-based planning by formulating a set of cooperative project planning problems. Each problem presents the LLM with a dialogue among participants discussing a project, where relevant temporal constraints are implicitly or explicitly mentioned. To comprehensively assess temporal reasoning, we introduce two categories of problems: \textit{short} and \textit{long}, where \textit{short} problems involve planning for projects completed within 1 day, and \textit{long} problems span up to 7 days.
Fig.~\ref{fig:data_example} shows an example for each category, and we summarize the types of involved temporal constraints in Table~\ref{tab:temporal_constraints}. 
Although all problems are based on common temporal constraints, due to the different temporal scope, the expressions of temporal constraints vary, and this requires models to comprehend diverse temporal concepts.
\textit{Short} problems mainly test models' temporal reasoning abilities including time zone interpretation and clock-time computation, while \textit{long} problems focus on weekday understanding and  calculations involving dates. 
\subsection{Benchmark Construction}
\label{sec: benchmark construction}
\begin{figure*}[htbp]
\centering
    \includegraphics[width=\textwidth]{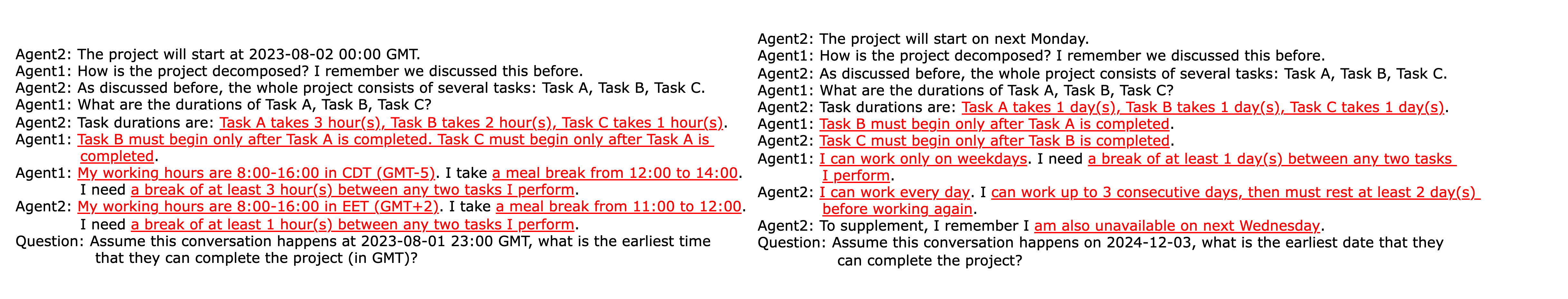}
\caption{Problem prototype examples. \textbf{Left}: prototype of the \textit{short} problem in Fig. \ref{fig:data_example} (left). \textbf{Right}: prototype of the \textit{long} problem in Fig. \ref{fig:data_example} (right). The red underlined contents correspond to the temporal constraints.}
\label{fig:prototype}
\vspace{-10pt}
\end{figure*}

We construct our benchmark using the following pipeline. First, we generate problem prototypes based on manually written constraint templates. Next, we create diverse project scenarios across different domains. Each prototype is then paired with a project scene to form a fully concretized problem. Finally, we perform a quality check on all concretized problems, manually correcting any that do not meet quality standards to ensure the overall reliability of the benchmark. In total, TCP contains 600 data instances, evenly split between the \textit{short} and \textit{long} categories. See Fig.~\ref{fig:construction} for an overview.
\begin{table}[htbp]
\centering
\resizebox{0.9\columnwidth}{!}{
\begin{tabular}{lc}
\toprule
\textbf{Dependency Type} & \textbf{Explanation} \\
\midrule
A $\rightarrow $ B, A $\rightarrow $ C & Task A should finish before the \\
& start of Task B and C. \\
\midrule
A $\rightarrow $ B, B $\rightarrow $ C & Task B starts after Task A finishes; \\
& Task C starts after Task B finishes. \\
\midrule
A $\rightarrow $ C, B $\rightarrow $ C & Task A and B should finish \\
& before the start of Task C.\\
\bottomrule
\end{tabular}
}
\caption{Dependency types.}
\label{tab:dependency}
\vspace{-10pt}
\end{table}
\paragraph{Problem Prototype Generation.}
For each problem category (\textit{short} and \textit{long}), we first manually implement a Python script, \texttt{generator.py}, to produce a set of problem prototypes along with their answers. Each prototype is based on a human-written natural language template and includes a sampled set of temporal constraints from Table~\ref{tab:temporal_constraints}. To ensure answer correctness, the script incorporates an exhaustive search algorithm during generation. Each prototype is then formulated as a dialogue between participants, a question about the optimal plan, and a verified ground truth answer. 
We generate 300 prototypes for each problem category, each involving two participants and three tasks as part of a cooperative project planning problem. Motivated by the findings of \citet{DBLP:conf/icml/LinMHY0P24}, which show that recent LLMs struggle with asynchronous planning, we introduce three distinct dependency types (listed in Table~\ref{tab:dependency}) and assign them evenly across prototypes to better analyze their impact. Fig.~\ref{fig:prototype} shows an example prototype for each category.
\paragraph{Project Scenario Generation.}
Since problem prototypes are generated from predefined templates, they lack linguistic variety and contextual richness. Tasks in prototypes are denoted using generic labels (e.g., Task A) and lack descriptive content, making the prototypes less reflective of real-world planning scenarios, which typically feature concrete task names and detailed descriptions. To address this, we select 10 common real-world domains—Computer Science, Business Administration, Mechanical Engineering, Biology, Economics, Psychology, Political Science, Electrical Engineering, Communications/Media Studies, and Nursing/Health Sciences—and manually create 3 distinct project scenarios per domain, each corresponding to a different dependency type. These domain-specific scenarios introduce both content diversity and broader topic coverage, while also ensuring a balanced distribution of dependency types. In total, this yields 30 initial (round-0) project scenarios for each problem category.

Each problem category (e.g., \textit{short} or \textit{long}) requires different project scenarios to ensure temporal feasibility. After creating the initial set of seed scenarios, we use an LLM, Gemini 2.5 Pro, to expand the scenario pool through 9 additional rounds of generation. In each round $r$, 30 new project scenarios are generated per problem category. 
For each new scenario, the LLM is prompted to draw inspiration from the round $r{-}1$ scenario within the same domain and dependency type, while avoiding close resemblance to those from rounds $0$ to $r{-}2$.
(see Fig.~\ref{fig:prompt_scenario_short} and~\ref{fig:prompt_scenario_long} in the Appendix for our prompts). Each generated scenario is reviewed by a human expert, who requests regeneration if the scenario is deemed low-quality. This process continues until all generated scenarios are approved.
As a result, each problem category contains 300 project scenarios evenly distributed across dependency types. Since we apply this process separately to both \textit{short} and \textit{long} categories, our benchmark includes a total of 600 high-quality, diverse project scenarios. 

\paragraph{Problem Concretization.}
With both prototypes and project scenarios prepared, we proceed to concretize the final benchmark problems for TCP. During this process, we associate each prototype with its corresponding dependency type and retrieve all project scenarios sharing that type. One scenario is randomly sampled and paired with the prototype for concretization. We then prompt Gemini 2.5 Pro to generate the final problem instance, using the selected prototype and project scenario as inspiration. The prompt instructs the model to enrich the dialogue with contextualized project details and to include logical justifications for each temporal constraint present in the prototype. Additionally, we ask the model to replace placeholder participant names (e.g., Agent 1) with realistic names. Since the first two stages maintain an even distribution of dependency types and domains, this balance naturally carries over into the concretized benchmark data. We provide the prompts in Fig.~\ref{fig:concretization_prompt_short} and~\ref{fig:concretization_prompt_long} in the Appendix.

\paragraph{Final Quality Check.}
Since we rely on an LLM for problem concretization, there remains a risk of undesired outputs such as altered temporal constraints or logical inconsistencies in the imagined context, even though we guide the model’s behavior through carefully designed prompts. To mitigate this, we implement a two-step quality control.

The first step, referred to as the hard check, is a symbolic verification carried out by a hard-coded Python script. When generating problem prototypes, we store their temporal constraints as structured metadata in a JSON object. During concretization, the LLM is also instructed to rewrite the placeholder names for participants and tasks in the metadata alongside the generated dialogue. We assume that if the model makes errors in metadata rewriting, it is highly likely to introduce inconsistencies in the dialogue or the question as well. The hard check ensures that the metadata remains consistent and correctly mapped. In our benchmark construction, Gemini 2.5 Pro successfully passed the hard check for all problems, indicating a high level of reliability in metadata handling. We believe the success rate is closely tied to the model’s capability and that this step offers an efficient way to detect failures in concretization.

Inspired by \citet{DBLP:conf/iclr/MiaoTR24}, the second step, the soft check, evaluates semantic consistency between the prototype and its concretized version. 
For each instance, we prompt Gemini 2.5 Pro (see Fig.~\ref{fig:soft_check_prompt} in the Appendix) to assess whether the prototype and its concretized problem represent the same scheduling scenario and question. The model is also asked to provide a justification. 
In total, we find 43 out of 600 problems fail the soft check, each accompanied by a reasonable justification. 
We review the model’s justifications and revise the failed cases manually.

To validate the quality of our generated data, 
we recruit five human annotators and assign each of them 10 randomly sampled, non-overlapping TCP problems. Annotators are given only the dialogues and questions and are allowed unlimited time to answer. 
For all sampled problems, the ground truth labels in the TCP benchmark match the answers provided by human annotators, confirming the reliability of the generated problems and the effectiveness of our quality assurance pipeline. We provide details of human annotation in Appendix \ref{app:humananno}, including guidelines for annotators and their background.

\begin{table}[!tb]
\centering
\resizebox{\columnwidth}{!}{
\begin{tabular}{@{}lccc@{}}
\toprule
\textbf{Models \& Methods} & \textbf{Short} & \textbf{Long} & \textbf{Overall} \\
\midrule
\rowcolor{gray!20} \multicolumn{4}{c}{\textbf{Naive Baseline}} \\
Random & 4.16 & 14.29 & 9.23  \\
\rowcolor{gray!20} \multicolumn{4}{c}{\textbf{Standard LLMs}} \\
Gemma 3 12B   &    0.67 &  1.00 &      0.83 \\
Gemma 3 27B   &    1.33 & 11.00 &      6.17 \\
LLaMA 3.1 8B  &    1.00 &  6.67 &      3.83 \\
LLaMA 3.1 70B &    1.67 &  3.67 &      2.67 \\
LLaMA 3.3 70B &    1.67 &  6.67 &      4.17 \\
Qwen3 8B     &    3.67 & 10.33 &      7.00 \\
Qwen3 14B    &    1.67 & 10.33 &      6.00 \\
 Gemini 2.0 Flash Lite & 3.00   & 10.00  & 6.50            \\
 Gemini 2.0 Flash      & 2.33   & 7.33   & 4.83            \\
 GPT-4.1 mini          & 10.33  & 9.00   & 9.67            \\
 GPT-4.1               & 10.00  & 10.33  & 10.17           \\
\midrule
\rowcolor{gray!20} \multicolumn{4}{c}{\textbf{Standard LLMs + Chain-of-Thought}} \\
Gemma 3 12B   &    7.44 &  7.33 &      7.39 \\
Gemma 3 27B   &    2.56 & 10.22 &      6.39 \\
LLaMA 3.1 8B  &    1.78 &  7.22 &      4.50 \\
LLaMA 3.1 70B &    6.44 & 11.78 &      9.11 \\
LLaMA 3.3 70B &   13.00 & 16.11 &     14.56 \\
Qwen3 8B     &    7.67 & 13.67 &     10.67 \\
Qwen3 14B    &   14.00 & 23.00 &     18.50 \\
Gemini 2.0 Flash Lite & 17.33 & 15.00 & 16.17              \\
Gemini 2.0 Flash & 11.33 & 20.00 & 15.67                      \\
GPT-4.1 mini & 40.33 & 62.00 & 51.17                      \\
GPT-4.1 & 50.33 & 65.00 & 57.67                            \\
\midrule
\rowcolor{gray!20} \multicolumn{4}{c}{\textbf{Reasoning LLMs}} \\
Qwen3 8B  &   21.56 & 42.67 &     32.11 \\
Qwen3 14B &   58.89 & 53.89 &     56.39 \\
Gemini 2.5 Flash & 45.33 & 78.67 & 61.98                \\
Gemini 2.5 Pro & 57.11 & 85.63 & 71.37                   \\
o4-mini & 76.33 & 86.44 & 81.39                      \\
\bottomrule
\end{tabular}
}
\caption{Benchmark results on TCP. We provide additional results in Table \ref{tab:main result complete}, including more Gemma and Qwen models of smaller sizes.}
\label{tab:main result}
\end{table}
\section{Experiments}
\subsection{Experimental Setting}
We benchmark two types of LLMs: (1) standard LLMs, including open-source models such as Gemma 3 \citep{DBLP:journals/corr/abs-2503-19786}, Llama 3 \citep{DBLP:journals/corr/abs-2407-21783}, and Qwen3 \citep{qwen3}, as well as proprietary models such as Gemini 1.5 Pro \citep{DBLP:journals/corr/abs-2403-05530}, Gemini 2.0 Flash Lite\footnote{All Gemini models are from: https://ai.google.dev/gemini-api/docs/models}, Gemini 2.0 Flash, GPT-4.1 mini\footnote{All OpenAI models are from: https://platform.openai.com\\/docs/overview}, and GPT-4.1; (2) reasoning LLMs, including the open-source Qwen3 with thinking enabled, and proprietary models such as Gemini 2.5 Flash, Gemini 2.5 Pro, and o4-mini. For completeness, we also include a naive baseline named Random by randomly selecting one answer from the potential answer candidates (choosing one of 24 hours and one of 7 days for \textit{short} and \textit{long} problems, respectively). We employ accuracy as evaluation metric. For standard LLMs, we report accuracies using greedy decoding, while for reasoning models, we run each of them 3 times with default configuration and report the average numbers\footnote{Reasoning models tend to perform worse under greedy decoding \cite{DBLP:journals/corr/abs-2501-12948,qwen3}.}. We also benchmark standard LLMs when they are coupled with Chain-of-Thought (CoT) prompting \citep{DBLP:conf/nips/Wei0SBIXCLZ22}. We provide more implementation details in Appendix \ref{app:implementations}.
\subsection{Benchmark Results}
We present the benchmark results in Table \ref{tab:main result} and highlight several key findings. (1) Our benchmark is highly challenging. Even the latest and strongest reasoning models such as o4-mini and Gemini 2.5 Pro cannot solve TCP problems optimally (with overall accuracy 81.39 and 71.37, respectively). Notably, many standard LLMs are outperformed by a naive random guessing baseline, even when prompted to produce CoTs. (2) Reasoning models significantly outperform standard LLMs on TCP. Generating CoTs also substantially improves the performance of standard models, indicating that test-time scaling is beneficial for solving TCP. (3) The performance gap between reasoning models and standard models, even with CoTs, suggests that TCP demands complex reasoning capabilities, further validating the benchmark’s difficulty. 
(4) Within model families evaluated under the same settings, we observe that larger model sizes generally lead to better TCP performance, particularly for reasoning models such as Qwen3.

\subsection{Further Analysis and Findings}

\begin{figure*}[t]
    \centering
    \includegraphics[width=\textwidth]{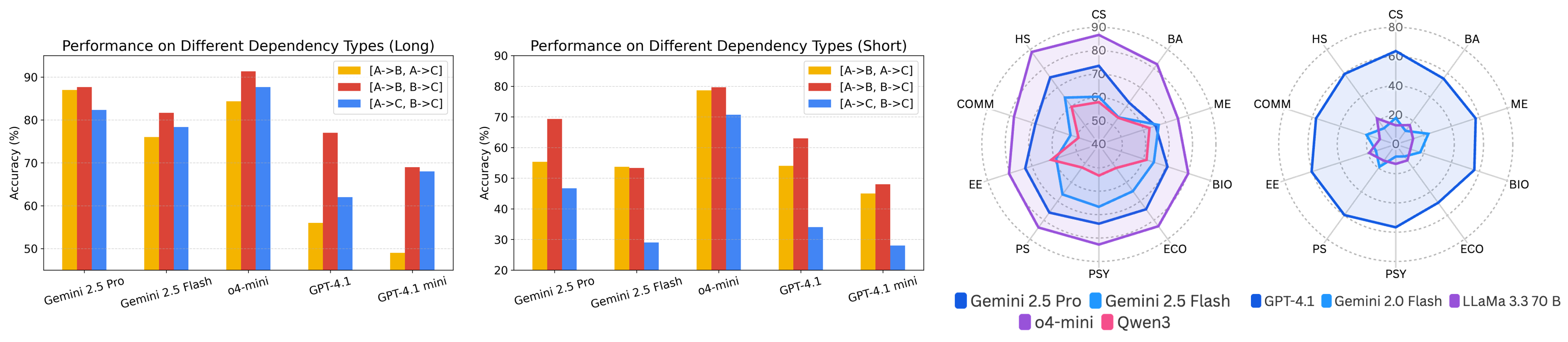}
    \caption{Performance on different dependency types for \textit{short} and \textit{long} problems (\textbf{left}) and  on different domains for reasoning and standard LLMs w. CoT prompting (\textbf{right}). CS = Computer Science,
    BA = Business Administration,
    ME = Mechanical Engineering,
    BIO = Biology,
    ECO = Economics,
    PSY = Psychology,
    PS = Political Science,
    EE = Electrical Engineering,
    COMM = Communications/Media Studies,
    HS = Nursing/Health Sciences.}
    \label{fig:combined_analysis}
\end{figure*}

\paragraph{Performance on Different Dependency Types.}
We analyze the influence of dependency types on the three strongest reasoning models: o4-mini, Gemini 2.5 Pro, and Gemini 2.5 Flash, as well as the two best-performing standard LLMs with CoT prompting: GPT-4.1 and GPT-4.1 mini (Fig.~\ref{fig:combined_analysis}). Across all models, we observe that sequential task dependencies are consistently easier to handle. In contrast, when problems require asynchronous planning, where tasks can be assigned to both participants to be carried out simultaneously, none of the models perform comparably. This suggests that even the most advanced models lack the ability to effectively plan asynchronously.  

\paragraph{Performance on Different Domains.}
We show domain-wise performance across different model families in Fig.~\ref{fig:combined_analysis}. Among the reasoning models, o4-mini demonstrates strong overall robustness, with slightly lower performance in Mechanical Engineering and Communications/Media Studies. Gemini 2.5 Pro also performs well overall, except in Mechanical Engineering and Business Administration. Interestingly, Gemini 2.5 Flash and Qwen3 exhibit similar strengths and weaknesses across various domains, although Gemini consistently outperforms Qwen3 in the remaining domains.
For standard LLMs with CoT prompting, GPT-4.1 from OpenAI remains the most robust across domains. LLaMA 3.3 70B and Gemini 2.0 Flash show varying performance, each excelling in some domains while underperforming in others. We conjecture that these domain-specific discrepancies may stem from differences in training data distribution and training strategies. In this regard, models from OpenAI appear to be the most well-rounded.

\begin{figure}[t]
    \centering
    \begin{subfigure}[t]{\columnwidth}
        \centering
        \includegraphics[width=\linewidth]{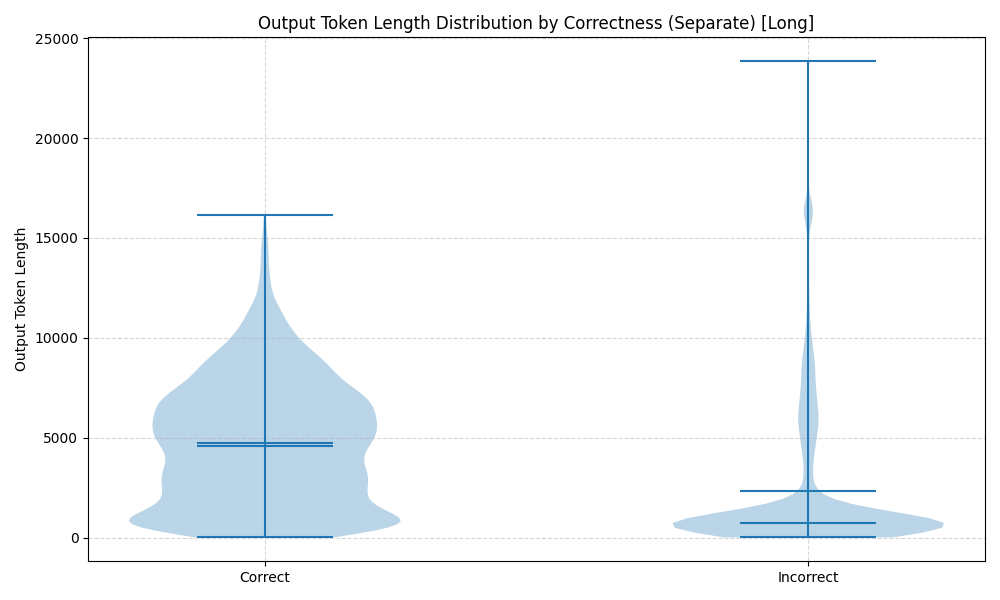}
        \caption{Long problems.}
        \label{fig:violin-long}
    \end{subfigure}
    
    \vspace{2mm} 

    \begin{subfigure}[t]{\columnwidth}
        \centering
        \includegraphics[width=\linewidth]{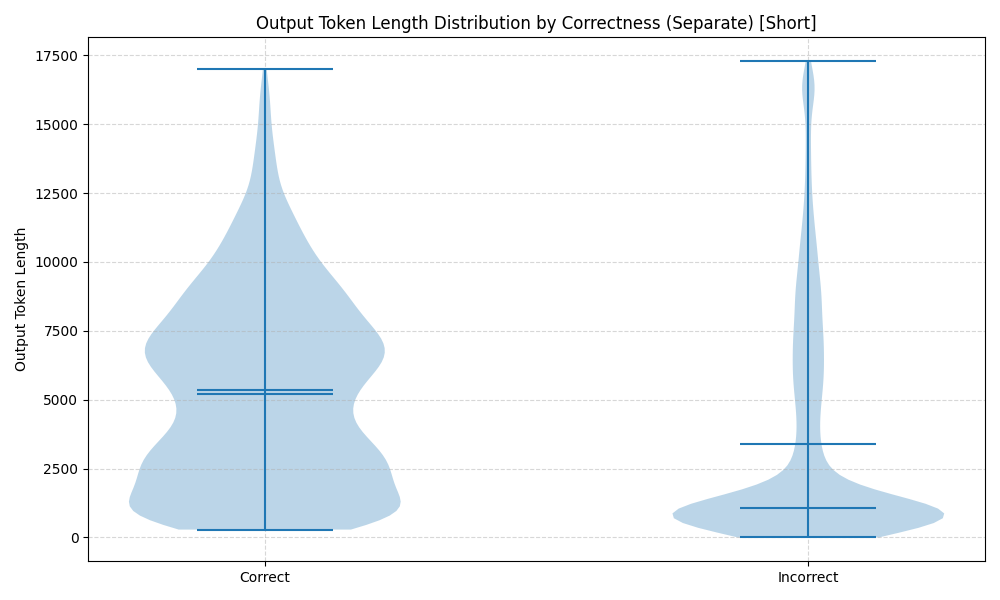}
        \caption{Short problems.}
        \label{fig:violin-short}
    \end{subfigure}
    
    \caption{Distribution of output lengths (number of tokens) for correct and incorrect predictions. Longer outputs tend to lead to correct answers and shorter outputs tend to correlate with incorrect predictions.}
    \label{fig:violin-combined}
    \vspace{-10pt}
\end{figure}

\paragraph{Performance on Varied Output Length.}
We analyze the relationship between output length, measured by the number of output tokens including the full output traces, and task accuracy. Specifically, we compare the output length distributions for correct and incorrect predictions across both the \textit{short} and \textit{long} problem categories. Fig. \ref{fig:violin-combined} shows violin plots of output token lengths. Incorrect predictions are strongly skewed towards shorter outputs, often clustered near zero, while correct predictions display a broader and more balanced distribution. The median and upper quartile lengths are substantially higher for correct responses, indicating that successful completions are typically associated with more extended reasoning. These results suggest that shorter outputs often reflect insufficient reasoning effort and highlight the importance of sustained reasoning in solving temporally constrained problems accurately.

\begin{figure}[t]
    \centering
    \includegraphics[width=\columnwidth, height=1.65\columnwidth]
    {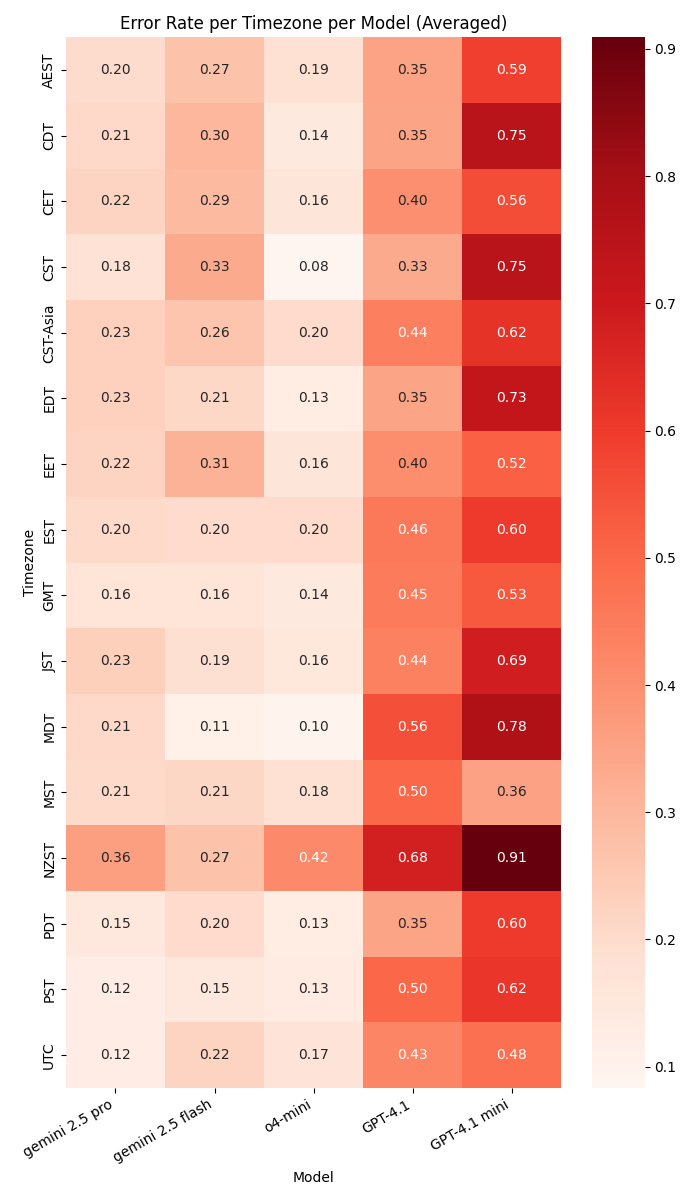}
    \caption{
    Error rates across time zones for strongest models.
    }
    \label{fig:error_rate_heatmap}
    \vspace{-10pt}
\end{figure}

\paragraph{Performance on Different Time Zones.}
We evaluate time zone-specific performance by selecting o4-mini, Gemini 2.5 Pro/Flash, and GPT-4.1/4.1 mini, and tracking their errors across time zones. If a problem references a particular time zone and a model fails to solve it, the error is attributed to that zone.
Although the heatmap in Fig.~\ref{fig:error_rate_heatmap} reveals some variation in error rates across time zones, a broader trend emerges: LLMs consistently struggle with time zone reasoning. While certain zones, such as NZST, exhibit substantially higher error rates, no single zone consistently accounts for the most errors across all models. One possible reason for NZST’s poor performance might be its relative rarity in training data compared to more frequently referenced zones like PST or GMT. Overall, many models show notable error rates across a wide range of time zones. This underscores the need for more robust temporal reasoning capabilities in LLMs. As discussed in Appendix~\ref{app:timezone}, models often rely on shallow heuristics that misinterpret time zone differences or overlook global temporal constraints, leading to persistent and compounding scheduling errors.

\paragraph{Reasoning LLMs vs. Standard LLMs + CoT.}
As shown in Table~\ref{tab:main result}, reasoning LLMs largely outperform standard LLMs prompted with CoT on TCP problems. To better understand this gap, we conduct two qualitative case studies.
The first case (Fig.~\ref{fig:qwen_comparison}) compares Qwen3 14B in thinking (reasoning) mode with its standard (non-thinking) mode on the same TCP problem. We find that the reasoning model frequently engages in iterative reflection (such as rethinking assumptions, exploring alternatives, and comparing outcomes), whereas with non-thinking mode, even prompted with CoT, model still tends to follow a fixed, linear path with little revision or exploration (as pointed out in Table~\ref{tab:reasoning_mode_impact_qwen}).
In the second case (Fig.~\ref{fig:gemini_gpt_comparison}), we contrast Gemini 2.5 Flash and GPT-4.1 mini. We find that Gemini adopts a structured, hierarchical reasoning layout. When solving a TCP problem, it typically decomposes the problem by task or time and anchors its reasoning around these subproblems. By contrast, GPT-4.1 mini + CoT often produces fragmented, ad hoc reasoning that lacks global consistency, leading to errors on complex, multi-constraint tasks (Table~\ref{tab:reasoning_mode_impact_gemini_gpt}). Together, these cases suggest that solving TCP problems requires not just temporal understanding but also the ability to coordinate reasoning across multiple interdependent constraints in a structured and reflective manner.

\section{Conclusion}
We present TCP, a benchmark for evaluating temporal constraint-based planning in dialogue-based scenarios. Our results show that even the strongest LLMs struggle, particularly with asynchronous planning and time zone reasoning. These challenges are often linked to shallow or fragmented reasoning, as evidenced by the correlation between longer, more structured outputs and higher accuracy. Notably, reasoning LLMs consistently outperform CoT-prompted standard LLMs on TCP by adopting more reflective and modular strategies that better handle complex constraint interactions. This contrast highlights the importance of not just surface-level temporal understanding, but also of structured reasoning processes. TCP thus exposes critical gaps in current models and offers a focused testbed for advancing their temporal planning capabilities in the future.


\section*{Acknowledgement}
Zifeng Ding, Zhangdie Yuan and Andreas Vlachos are supported by the ERC grant AVeriTeC (GA 865958).
Andreas Vlachos is further supported by the DARPA program SciFy. Fangru Lin is supported by Clarendon and Jason Hu studentship.

\section*{Limitations}
While \textsc{TCP} offers a comprehensive benchmark for temporal constraint-based planning, it has several limitations. First, although the scenarios are naturalistic, they are generated by LLMs and may not fully capture the ambiguity and messiness of real-world planning dialogues. Second, we fix the number of agents and tasks (2 agents and 3 tasks) across all instances. This is a deliberate design choice to avoid benchmark difficulty being dominated by combinatorial explosion, allowing us to focus on evaluating planning over temporal constraints. However, this simplification limits our ability to assess model performance on more complex, large-scale planning problems. Third, models are evaluated without task-specific fine-tuning, which may underestimate their full potential.

\section*{Ethic Considerations}
This work does not involve human subject data or sensitive personal information. However, we note that LLMs’ planning capabilities, while promising, can be misleading if over-trusted. Poor temporal reasoning, especially in high-stakes domains such as healthcare or logistics, could result in harmful decisions if deployed without oversight. Our benchmark is intended for research purposes only and should not be construed as validating real-world deployment readiness. We encourage future work to pair model advances with interpretability and robustness checks, particularly when applying temporal planning models in safety-critical settings.

\bibliography{custom}

\appendix
\section{Implementation Details}
\label{app:implementations}
All experiments of open-source models are run with PyTorch \citep{DBLP:conf/nips/PaszkeGMLBCKLGA19} and vLLM \citep{DBLP:conf/sosp/KwonLZ0ZY0ZS23} on a machine with 96 CPU cores and 4 $\times$ Nvidia H100 GPU each with 94GB. For Gemini models, we run them with Google's python-genai\footnote{https://github.com/googleapis/python-genai} API. For OpenAI models, we run with OpenAI's openai-python\footnote{https://github.com/openai/openai-python} API.

We also list the detailed model versions used in our experiments. 
\paragraph{Open-Source Models.} We use official repositories of Gemma 3 family, including Gemma 3 1B it\footnote{https://huggingface.co/google/gemma-3-1b-it}, Gemma 3 4B it\footnote{https://huggingface.co/google/gemma-3-4b-it}, Gemma 3 12B it\footnote{https://huggingface.co/google/gemma-3-12b-it} and Gemma 3 27B it\footnote{https://huggingface.co/google/gemma-3-27b-it}. 
We use official repositories of LLaMA famility, including LLaMA 3.1 8B Instruct\footnote{https://huggingface.co/meta-llama/Llama-3.1-8B-Instruct}, LLaMA 3.1 70B Instruct\footnote{https://huggingface.co/meta-llama/Llama-3.1-70B-Instruct} and LLaMA 3.3 70B Instruct\footnote{https://huggingface.co/meta-llama/Llama-3.3-70B-Instruct}
We use official repositories of Qwen3 family, including Qwen3 4B\footnote{https://huggingface.co/Qwen/Qwen3-4B}, Qwen3 8B\footnote{https://huggingface.co/Qwen/Qwen3-8B}, Qwen3 14B\footnote{https://huggingface.co/Qwen/Qwen3-14B} and Qwen3 32B\footnote{https://huggingface.co/Qwen/Qwen3-32B}.

\paragraph{Proprietary Models.}
For Gemini models, we use: gemini-1.5-pro-002, gemini-2.0-flash-lite-001, gemini-2.0-flash-001, gemini-2.5-flash-preview-04-17 and gemini-2.5-pro-preview-05-06.
For OpenAI models, we use: gpt-4.1-mini-2025-04-14, gpt-4.1-2025-04-14 and o4-mini-2025-04-16.
\section{Complete Benchmark Results}
We present the complete performance results across all models and settings in Table \ref{tab:main result complete}.

\begin{table}[htbp]
\centering
\resizebox{\columnwidth}{!}{
\begin{tabular}{@{}lccc@{}}
\toprule
\textbf{Models \& Methods} & \textbf{Short} & \textbf{Long} & \textbf{Overall} \\
\midrule
\rowcolor{gray!20} \multicolumn{4}{c}{\textbf{Naive Baseline}} \\
Random & 4.16 & 14.29 & 9.23  \\
\rowcolor{gray!20} \multicolumn{4}{c}{\textbf{Standard LLMs}} \\
Gemma 3 1B    &    0.00 &  0.00 &      0.00 \\
Gemma 3 4B    &    1.33 &  4.33 &      2.83 \\
Gemma 3 12B   &    0.67 &  1.00 &      0.83 \\
Gemma 3 27B   &    1.33 & 11.00 &      6.17 \\
LLaMA 3.1 8B  &    1.00 &  6.67 &      3.83 \\
LLaMA 3.1 70B &    1.67 &  3.67 &      2.67 \\
LLaMA 3.3 70B &    1.67 &  6.67 &      4.17 \\
Qwen3 4B     &    4.33 &  7.67 &      6.00 \\
Qwen3 8B     &    3.67 & 10.33 &      7.00 \\
Qwen3 14B    &    1.67 & 10.33 &      6.00 \\
 Gemini 1.5 Pro        & 2.00   & 8.33   & 4.67            \\
 Gemini 2.0 Flash Lite & 3.00   & 10.00  & 6.50            \\
 Gemini 2.0 Flash      & 2.33   & 7.33   & 4.83            \\
 GPT-4.1 mini          & 10.33  & 9.00   & 9.67            \\
 GPT-4.1               & 10.00  & 10.33  & 10.17           \\
\midrule
\rowcolor{gray!20} \multicolumn{4}{c}{\textbf{Standard LLMs + Chain-of-Thought}} \\
Gemma 3 1B    &    0.22 &  0.89 &      0.56 \\
Gemma 3 4B    &    0.11 &  1.00 &      0.56 \\
Gemma 3 12B   &    7.44 &  7.33 &      7.39 \\
Gemma 3 27B   &    2.56 & 10.22 &      6.39 \\
LLaMA 3.1 8B  &    1.78 &  7.22 &      4.50 \\
LLaMA 3.1 70B &    6.44 & 11.78 &      9.11 \\
LLaMA 3.3 70B &   13.00 & 16.11 &     14.56 \\
Qwen3 4B     &    8.33 & 10.67 &      9.50 \\
Qwen3 8B     &    7.67 & 13.67 &     10.67 \\
Qwen3 14B    &   14.00 & 23.00 &     18.50 \\
Gemini 1.5 Pro & 12.33 & 14.33 & 13.33                     \\
Gemini 2.0 Flash Lite & 17.33 & 15.00 & 16.17              \\
Gemini 2.0 Flash & 11.33 & 20.00 & 15.67                      \\
GPT-4.1 mini & 40.33 & 62.00 & 51.17                      \\
GPT-4.1 & 50.33 & 65.00 & 57.67                            \\
\midrule
\rowcolor{gray!20} \multicolumn{4}{c}{\textbf{Reasoning LLMs}} \\
Qwen3 4B  &   28.11 & 36.00 &     32.06 \\
Qwen3 8B  &   21.56 & 42.67 &     32.11 \\
Qwen3 14B &   58.89 & 53.89 &     56.39 \\
Gemini 2.5 Flash & 45.33 & 78.67 & 61.98                \\
Gemini 2.5 Pro & 57.11 & 85.63 & 71.37                   \\
o4-mini & 76.33 & 86.44 & 81.39                      \\
\bottomrule
\end{tabular}
}

\caption{Complete benchmark results on TCP. }
\label{tab:main result complete}
\end{table}

\section{Human Annotation Details}
\label{app:humananno}
\begin{figure*}
    \includegraphics[width=\textwidth]{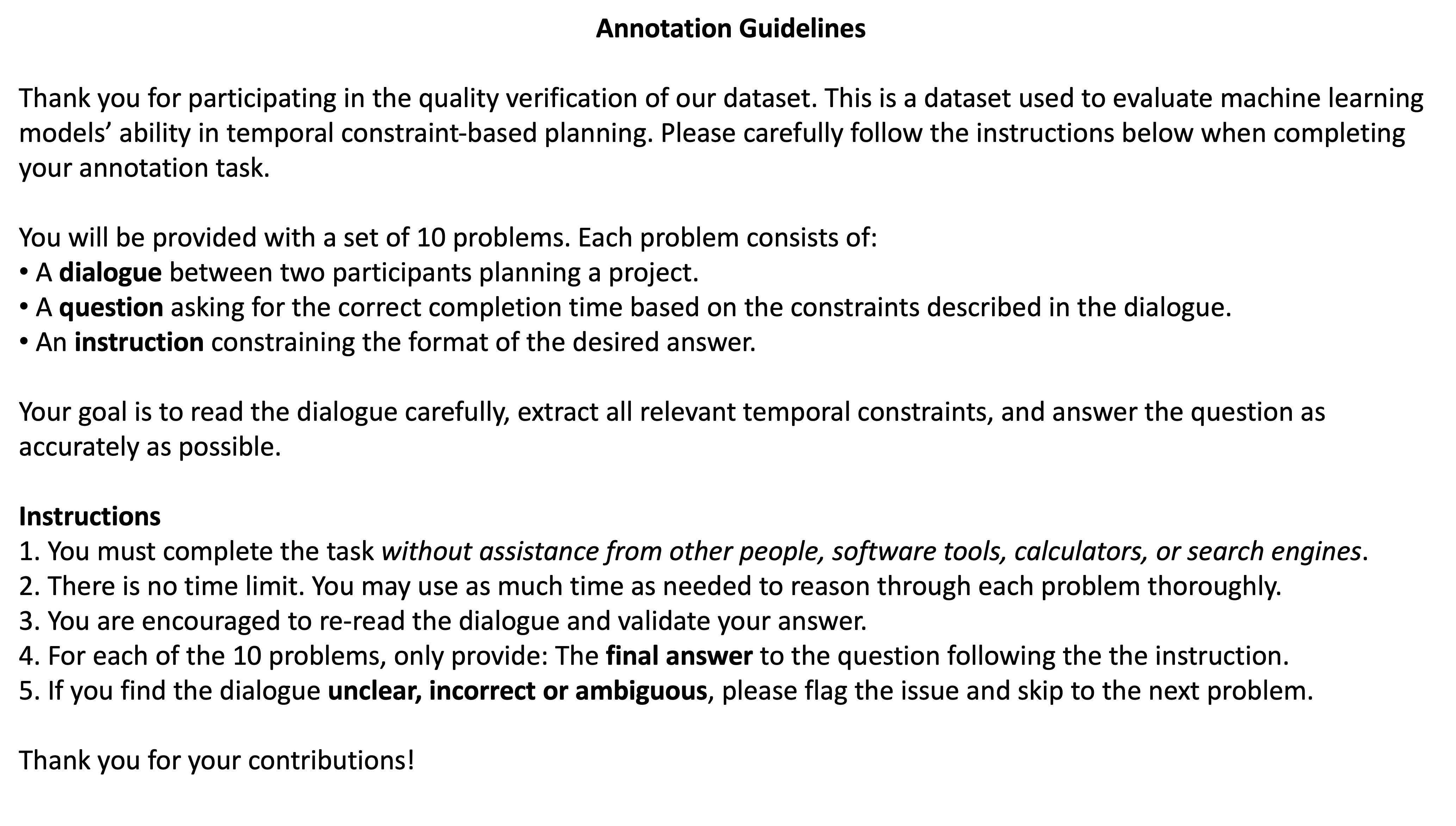}
    \caption{Annotation guideline for quality verification.}
    \label{fig:annoguideline}
\end{figure*}

We recruit five human annotators to evaluate the quality of our benchmark. All annotators are PhD students in Computer Science with at least full professional proficiency in English. All of them consent our usage of their data. The annotation guidelines are provided in Fig.~\ref{fig:annoguideline}.
\section{Performance on Problem Prototypes}
\label{sec:prototype_performance}
We evaluate model performance on abstract problem prototypes to isolate their ability to reason over temporal constraints without the added complexity of naturalistic language. As shown in Table \ref{tab:prototype result}, all models (Gemini 2.5 Flash, Gemini 2.5 Pro, and o4-mini) perform notably better on prototypes than on their concretized counterparts. We focus this analysis on the strongest models from our main evaluation to ensure that observed errors reflect reasoning limitations rather than deficiencies in basic model capacity. The consistent performance gap suggests that linguistic surface form introduces new challenges that disrupt otherwise successful reasoning. In the following analysis, we examine two prominent error types to better understand why concretized instances are harder.
\begin{table}[htbp]
\centering
\resizebox{\columnwidth}{!}{
\begin{tabular}{@{}lccc@{}}
\toprule
\textbf{Models \& Methods} & \textbf{Short} & \textbf{Long} & \textbf{Overall} \\
\midrule
\rowcolor{gray!20} \multicolumn{4}{c}{\textbf{Reasoning LLMs}} \\
Gemini 2.5 Flash & 53.67 & 81.33 & 67.50 \\
Gemini 2.5 Pro & 60.89 & 85.67 & 73.28   \\
o4-mini & 79.78 & 94.56 & 87.17 \\

\bottomrule
\end{tabular}
}
\caption{Results on problem prototypes.}
\label{tab:prototype result}
\end{table}
\begin{table}[htbp]
\centering
\resizebox{\columnwidth}{!}{
\begin{tabular}{@{}lccc@{}}
\toprule
\textbf{Models \& Methods} & \textbf{Short} & \textbf{Long} & \textbf{Overall} \\
\midrule
\rowcolor{gray!20} \multicolumn{4}{c}{\textbf{Reasoning LLMs}} \\
Gemini 2.5 Flash & 47.40 & 73.50 & 60.45 \\
Gemini 2.5 Pro & 57.78 & 85.50 & 71.64   \\
o4-mini & 56.78 & 81.89 & 69.34 \\

\bottomrule
\end{tabular}
}
\caption{Results after providing time-related world knowledge.}
\label{tab:dual result}
\end{table}

\input{error_samples/error_samples_prototype_vs_concretizations}

\section{Time Zone Error Analysis}
\label{app:timezone}
\input{error_samples/error_samples_timezone}

\section{Output Trace Comparison between Reasoning LLMs and Standard LLMs + CoT}
\label{app:reasoning_vs_cot}
\input{reasoning_path_comparison/reasoning_vs_standard}

\section{Dual Interpretation of Temporal Reasoning}
Many temporal reasoning skills such as time zone conversion and date–weekday transformation can be interpreted in two ways: either as independent temporal reasoning abilities or as a combination of world knowledge and reasoning. For example, if a model knows what each time zone represents, it can apply mathematical reasoning to compute the relative difference between two zones and thereby answer time-related questions. In this work, following prior studies on temporal reasoning \citep{DBLP:conf/acl/Wang024}, we adopt the first interpretation and treat temporal reasoning as an independent ability.

As a supplement, drawing on the second interpretation of temporal reasoning, we conduct a separate analysis in which we provide additional contextual information to test whether time-related world knowledge can assist reasoning LLMs in time constraint-based planning. For \textit{long} problems, the prompt includes the weekday corresponding to the date mentioned in the question, whereas for \textit{short} problems, it includes the relative distances between time zones. The experimental results in Table~\ref{tab:dual result} indicate that even with such additional knowledge, model performance does not improve. We attribute this to the fact that the main challenge in time constraint-based planning lies in managing multiple interdependent temporal constraints, a difficulty that cannot be alleviated merely by supplying time-related knowledge.

\section{Prompts}
\label{app:prompts}
We provide in this section all the prompts that we used for data construction and model evaluation.
\input{prompt_short}
\begin{figure*}[htbp]
\centering
\noindent\fbox{%
\parbox{\textwidth}{%
\small
\textbf{Prompt for \textit{Long} Project Scenario Generation:}
\begin{quote}
You are a project designer. Your job is to invent a new, realistic, and creative project in the area of '\textbf{\{area\}}', inspired by the following example project. Your new project should:\\
- Have a similar number of tasks (with distinct, concrete names and descriptions).\\
- Preserve the same dependency structure between tasks (see below).\\
- Be original, not a copy or trivial rewording of the example or any previously generated project (if provided).\\
- Be plausible for a real team to execute in this domain.\\
- Each task in the project must be scheduled to finish within 1-3 days.\\
\\
Here are the previously generated projects. The new project MUST NOT copy or trivially reword any of them. Use them as negative examples to ensure originality:\\
\textbf{\{prev\_refs\}}\\

Example project (format):\\
\textbf{\{example\_json\}}\\
\\
---

Please output a new project as a JSON object in the EXACT SAME FORMAT as the example above, with all the following fields:\\
- project\_name: (str) The new project name\\
- tasks: (list of dict) Each with 'task\_name' and 'possible\_contents'\\
- dependencies: (list of [from\_task, to\_task])\\
- task\_name\_map: (dict) Mapping new task names to generic names (e.g., 'Task A', 'Task B', ...)\\
- dependencies\_map: (list of [from\_task\_generic\_name, to\_task\_generic\_name]), mapping each new task name to its corresponding generic task name (e.g., 'Task A', 'Task B', ...) from dependencies\\
\\
IMPORTANT: Only output the JSON object, no commentary or markdown. Maintain the same field names and structure as the example.
\end{quote}
}
}
\caption{Complete prompt for project scenario generation for \textit{long} problems. \textbf{\{area\}} denotes a domain, e.g., Computer Science. \textbf{\{prev\_refs\}} represents the generated project scenarios within the same domain and dependency type from rounds $0$ to $r-2$. \textbf{\{example\_json\}} is the project scenario within the same domain and dependency type generated from the previous round $r-1$.}
\label{fig:prompt_scenario_long}
\end{figure*}

\begin{figure*}[htbp]
\centering
\noindent\fbox{%
\parbox{\textwidth}{%
\small
\textbf{Prompt for Concretizing \textit{Short} Problems:}
\begin{quote}
You are given a scheduling problem in a JSON structure. Your task is to rewrite ("evolve") the scenario to make it realistic, concrete, and vivid, following these rules:

1. Invent a specific project scenario and ensure all tasks and dialogue are consistent with this scenario. Use the following project as inspiration: [\textbf{\{area\}} - \textbf{\{project\_name\}}]. The main tasks are: \\
\textbf{\{task\_descriptions\}} \\
Make the dialogue and scenario sound like a real project, using your imagination for context. \\
2. Make sure the person who asks the task durations will not be the same one who answers this question. Similarly, make sure the person who asks the project decomposition will not be the same one who answers this question. \\
3. Replace generic agent names (e.g., 'Agent1', 'Agent2') with realistic, human names. Use your imagination to invent suitable names; do not use a predefined list. \\
4. Replace generic task names (e.g., 'Task A', 'Task B', 'Task C') with highly specific, concrete real-world tasks that clearly belong to the invented scenario. Avoid vague or generic task names. The workload of each task should be consistent with the duration. The dependencies between concretized tasks should be consistent with the original dependencies between generic tasks. The dependencies between concretized tasks should be logical. \\
5. When changing the task and agent names, make sure to update the corresponding fields in the JSON: "tasks", "agents", "dependencies", and "dependency\_graph" so that the structure is consistent and all references match the new names. \\
6. For every agent's availability constraint mentioned in the dialogue (such as "I can work up to 3 consecutive hours, then must rest at least 1 hour(s) before working again." or "I need a break of at least 1 hour(s) between any two tasks I perform."), add a natural-sounding reason or justification. For example: "because I need to spend some time with my newly-born baby". The reason should be plausible and detailed and fit the context of the scenario. Note that the reason should be logical to happen in the period of rest time. \\
7. Except changing the generic names (e.g., 'Agent1', 'Agent2') to the real names in "agent\_constraints" and "agent\_constraints\_gmt", do not change the numbers in "agent\_constraints" and "agent\_constraints\_gmt". \\
8. You may freely paraphrase any sentence in the dialogue, as long as you do not alter the logic or the facts in the original sentence. You can also paraphrase the question, but do not change the core meaning. \\
9. Do NOT change any numbers, constraints, durations, or logic in the data. Use "hour" or "hours" to substitute "hour(s)". \\
10. The entire project must be scheduled to finish within a day. Make sure all dialogue and constraints are consistent with this total project duration. \\
11. Output the result as a JSON object with the same structure as the input, but with the scenario, dialogue, names, and relevant fields concretized. \\

IMPORTANT: Do not include \texttt{```json} at the beginning and \texttt{```} at the end of the output. Do not generate any comment. \\

Here is the original data:

\textbf{\{problem\_prototype\}}
\end{quote}
}
}
\caption{Complete prompt for concretizing \textit{short} problems. \textbf{\{area\}} denotes a domain, e.g., Computer Science. \textbf{\{project\_name\}} is the name of the sampled project scenario. \textbf{\{task\_descriptions\}} denotes the descriptions of different tasks in the sampled project scenario. Descriptions are generated together with each project scenario. \textbf{\{project\_prototype\}} represents project prototype.}
\label{fig:concretization_prompt_short}
\end{figure*}

\begin{figure*}[htbp]
\centering
\noindent\fbox{%
\parbox{\textwidth}{%
\small
\textbf{Prompt for Concretizing \textit{long} Problems:}
\begin{quote}
You are given a scheduling problem in a JSON structure. Your task is to rewrite ("evolve") the scenario to make it realistic, concrete, and vivid, following these rules:

1. Invent a specific project scenario and ensure all tasks and dialogue are consistent with this scenario. Use the following project as inspiration: [\textbf{\{area\}} - \textbf{\{project\_name\}}]. The main tasks are: \\
\textbf{\{task\_descriptions\}} \\
Make the dialogue and scenario sound like a real project, using your imagination for context. \\
2. Make sure the person who asks the task durations will not be the same one who answers this question. Similarly, make sure the person who asks the project decomposition will not be the same one who answers this question. \\
3. Replace generic agent names (e.g., 'Agent1', 'Agent2') with realistic, human names. Use your imagination to invent suitable names; do not use a predefined list. \\
4. Replace generic task names (e.g., 'Task A', 'Task B', 'Task C') with highly specific, concrete real-world tasks that clearly belong to the invented scenario. Avoid vague or generic task names. The workload of each task should be consistent with the duration. The dependencies between concretized tasks should be consistent with the original dependencies between generic tasks. The dependencies between concretized tasks should be logical. \\
5. When changing the task and agent names, make sure to update the corresponding fields in the JSON: "tasks", "agents", "dependencies", and "dependency\_graph" so that the structure is consistent and all references match the new names. \\
6. For every agent's availability constraint mentioned in the dialogue (such as "I can work up to 3 consecutive days, then must rest at least 1 day(s) before working again." or "I need a break of at least 1 day(s) between any two tasks I perform."), add a natural-sounding reason or justification. For example: "because I need to spend some time with my newly-born baby". The reason should be plausible and detailed and fit the context of the scenario. Note that the reason should be logical to happen in the period of rest time. \\
7. Except changing the generic names (e.g., 'Agent1', 'Agent2') to the real names in "agent\_unavailable\_dates", do not change the numbers in "agent\_unavailable\_dates". \\
8. You may freely paraphrase any sentence in the dialogue, as long as you do not alter the logic or the facts in the original sentence. You can also paraphrase the question, but do not change the core meaning. \\
9. Do NOT change any numbers, constraints, durations, or logic in the data. Use "day" or "days" to substitute "day(s)".\\
10. The entire project must be scheduled to finish within 7 days. Make sure all dialogue and constraints are consistent with this total project duration. \\
11. Output the result as a JSON object with the same structure as the input, but with the scenario, dialogue, names, and relevant fields concretized. \\

IMPORTANT: Do not include \texttt{```json} at the beginning and \texttt{```} at the end of the output. Do not generate any comment. \\

Here is the original data:

\textbf{\{problem\_prototype\}}
\end{quote}
}
}
\caption{Complete prompt for concretizing \textit{long} problems. \textbf{\{area\}} denotes a domain, e.g., Computer Science. \textbf{\{project\_name\}} is the name of the sampled project scenario. \textbf{\{task\_descriptions\}} denotes the descriptions of different tasks in the sampled project scenario. Descriptions are generated together with each project scenario. \textbf{\{project\_prototype\}} represents project prototype.}
\label{fig:concretization_prompt_long}
\end{figure*}

\begin{figure*}[t]
\centering
\noindent\fbox{%
\parbox{\textwidth}{%
\small
\textbf{Prompt for LLM-Based Soft Check:}
\begin{quote}
You are given two scheduling problems. Your job is to judge whether they represent the same scheduling scenario and question in essence.

Pay special attention to:\\
- Whether the dialogues are representing the same scenario (project structure, dependencies, constraints, etc.)\\
- Whether the questions are generally the same (even if task contents are different)\\

If they are essentially the same, answer YES and briefly explain why. If not, answer NO and explain the main differences. Wrap YES or NO in \textbackslash boxed\{\}. Always provide your answer before giving explanation.\\

---

[Evolved/Seed Dialogue]:\\
\textbf{\{problem\_dialogue\}}\\

[Evolved/Seed Question]: \textbf{\{problem\_question\}}\\

[Original Dialogue]:\\
\textbf{\{prototype\_dialogue\}}\\

[Original Question]: \textbf{\{prototype\_question\}}\\

---
\end{quote}
}
}
\caption{Complete prompt for the LLM-based soft check. \textbf{\{problem\_dialogue\}}/\textbf{\{problem\_question\}} and \textbf{\{prototype\_dialogue\}}/\textbf{\{prototype\_question\}} denote the dialogue/question of a concretized problem and its corresponding project prototype, respectively. }
\label{fig:soft_check_prompt}
\end{figure*}

\section{Additional Example Problems}
\begin{figure*}[htbp]
    \centering
    \includegraphics[width=\textwidth]{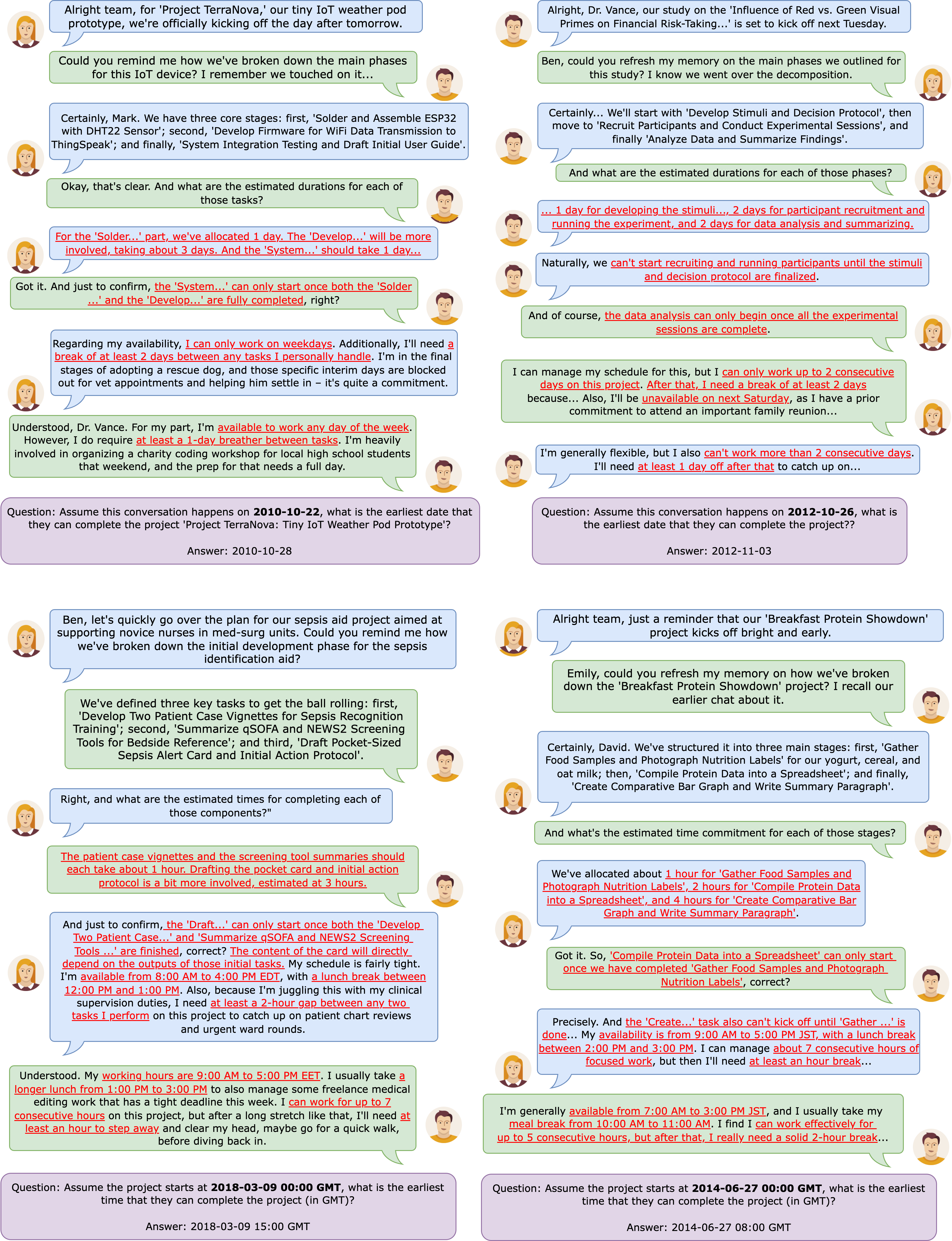}
    \caption{Additional example \textit{short} (\textbf{bottom}) and \textit{\textbf{long}} (top) problems in TCP. 
    }
    \label{fig:data_example_more}
    \vspace{-10pt}
\end{figure*}

We provide additional example problems in Fig.~\ref{fig:data_example_more}. The top two are \textit{long} problems, while the bottom two are \textit{short} problems. Specifically, the top-left problem corresponds to Electrical Engineering with dependency type A $\rightarrow$ C, B $\rightarrow$ C, the top-right to Psychology with dependency type A $\rightarrow$ B, B $\rightarrow$ C, the bottom-left to Biology with dependency type A $\rightarrow$ B, A $\rightarrow$ C, and the bottom-right to Nursing/Health Sciences with dependency type A $\rightarrow$ C, B $\rightarrow$ C.
\label{app: examples}

\end{document}

%% file: error_samples/error_samples_prototype_vs_concretizations.tex
\paragraph{Surface Form Impacts Temporal Reasoning.}
We explain this type of error with the example illustrated in Fig.~\ref{fig:surface_form_impact} and Table~\ref{tab:surface_form_impact}.
Although the prototype and concretized dialogues describe the exact same scenario, only the prototype resulted in the correct project completion date of \textbf{September 14}. In the prototype, the model correctly anchored task scheduling to the explicitly stated kickoff on \textbf{Saturday, September 11}, assigning Task A to that date and computing downstream task timing accordingly. However, in the more naturalistic version, the model erroneously began scheduling from \textbf{September 7}, the date of the conversation despite the same kickoff being mentioned. This caused it to predict an invalid completion date of \textbf{September 9}. The shift appears to result from increased surface complexity: while the prototype used structured, abstract language, the concretized version contained narrative elements, justifications for constraints, and conversational phrasing that likely distracted the model from the global temporal constraint. This highlights how even when the semantics are preserved, variation in surface form can significantly impair temporal reasoning. See Fig.~\ref{fig:surface_form_impact} and Table~\ref{tab:surface_form_impact} for details. These results suggest that models like Gemini 2.5 Flash are sensitive to surface cues and may deprioritize global temporal anchors when interpreting more naturalistic input.

\paragraph{Surface Form Reduces Reasoning Length.}
One of the most intriguing failure modes we observe arises from the gap between abstract prototypes and their concretized natural language counterparts. While the concretization introduces more realistic language and richer narrative cues, it paradoxically leads to \textit{shorter} reasoning chains. This is counterintuitive: concretized examples contain more linguistic surface variability, named entities, and natural phrasing, so we would expect them to require more, not less, reasoning to reach a correct answer. Yet, when evaluated with Gemini 2.5 Flash, we observe a consistent and statistically significant drop in reasoning length: prototype examples average 7395.12 tokens in the reasoning trace, while concretized versions average only 5992.83 tokens.
This reduction in token length is not merely cosmetic but also correlates with worse performance. In a representative example, the model correctly solves the abstract prototype by leveraging parallelism across agents and respecting all constraints, but fails on the concretized version by redundantly assigning all tasks to a single agent, resulting in delayed project completion. This suggests that the model allocates less cognitive effort to naturalistic inputs, possibly due to implicit assumptions about verbosity or prompt familiarity, leading to underdeveloped planning in more lifelike scenarios. See Fig.~\ref{fig:verbatim_comparison_colored} and Table~\ref{tab:token_length_impact} for example illustration.

\input{error_samples/prototype_vs_concretization_table}

\input{error_samples/prototype_vs_cibcretuzation_token_size_table}

%% file: error_samples/prototype_vs_concretization_table.tex
\begin{figure*}[htbp]
\centering
\noindent\fbox{%
\parbox{\textwidth}{%
\small
\textbf{Original Dialogue:}
\begin{quote}
Olivia: Great news, Liam! We're kicking off the RC monster truck shock absorber project this Saturday.\\
Liam: Sounds good, Olivia. Could you remind me of the main task breakdown we agreed on for this initial concept phase?\\
Olivia: Certainly, Liam. We'll start with \textbf{\textcolor{orange}{"Define Load \& Damping for RC Truck"}}, then move to \textbf{\textcolor{orange}{"Sketch Piston-Cylinder Assembly"}}, and finally \textbf{\textcolor{orange}{"Specify Spring \& Damping Fluid"}}.\\
Liam: `Define Load \& Damping' should take \textcolor{blue}{1 day}. `Sketch Piston-Cylinder Assembly' takes \textcolor{blue}{1 day}, and `Specify Spring \& Damping Fluid' takes \textcolor{blue}{2 days}.\\
Liam: \textbf{\textcolor{red}{"Sketch"}} and \textbf{\textcolor{red}{"Specify"}} both depend on the completion of \textbf{\textcolor{red}{"Define Load \& Damping"}}.\\
Liam: I can work any day, but after \textcolor{purple}{2 consecutive days}, I need a \textcolor{purple}{1-day rest}.\\
Olivia: I work only on weekdays, need \textcolor{teal}{1 day break between tasks}, and will be unavailable next \textcolor{teal}{Monday, Sept 13}.
\end{quote}
}
}
\caption{Concretized version of the RC shock absorber dialogue, annotated with task descriptions (\textcolor{orange}{task names}), durations (\textcolor{blue}{durations}), dependencies (\textcolor{red}{dependencies}), Liam’s constraints (\textcolor{purple}{work/rest}), and Olivia’s constraints (\textcolor{teal}{weekday-only, breaks, unavailability}).}
\label{fig:surface_form_impact}
\end{figure*}

\begin{table*}[htbp]
\centering
\small
\begin{tabular}{p{0.28\textwidth} p{0.32\textwidth} p{0.32\textwidth}}
\toprule
\textbf{Aspect} & \textbf{Prototype Prediction} & \textbf{Concretized Prediction} \\
\midrule
Project kickoff date interpretation & \textcolor{blue}{Correctly anchored to Saturday, Sept 11} & \textcolor{red}{Anchored to dialogue date, Sept 7} \\
Task scheduling & \textcolor{blue}{Task A on Sept 11; others follow respecting constraints} & \textcolor{red}{Task A on Sept 7; subsequent tasks conclude by Sept 9} \\
Final predicted completion date & \textbf{Sept 14 (correct)} & \textbf{Sept 9 (invalid)} \\
Constraint enforcement & \textcolor{blue}{Global project start constraint maintained} & \textcolor{red}{Overwritten by local availability reasoning} \\
Possible cause & Structured prompt helped retain constraints & Natural dialogue introduced distractors \\
\bottomrule
\end{tabular}
\caption{Despite identical constraints, the model adhered to the project start date in the abstract prototype but violated it in the concretized version.}
\label{tab:surface_form_impact}
\end{table*}

%% file: error_samples/prototype_vs_cibcretuzation_token_size_table.tex
\begin{figure*}[htbp]
\centering
\noindent\fbox{%
\parbox{\textwidth}{%
\small
\begin{tabular}{@{}p{0.48\textwidth}@{\hskip 0.02\textwidth}p{0.48\textwidth}@{}}
\textbf{Prototype Dialogue} & \textbf{Concretized Dialogue} \\
\toprule
Agent2: The project will start at 2014-12-17 00:00 GMT. & Michael: Alright Sarah, let's get this web series analysis project rolling. We need to wrap this initial phase up quickly, aiming to finish within the next two days. \\
Agent1: How is the project decomposed? I remember we discussed this before. & Sarah: Sounds good, Michael. Could you remind me how we're breaking down these initial steps for the audience reception part? I know we chatted about it. \\
Agent2: As discussed before, the whole project consists of several tasks: \textcolor{orange}{Task A}, \textcolor{orange}{Task B}, \textcolor{orange}{Task C}. & Michael: Sure thing. For this first stage, we have three main tasks: First, we need to \textcolor{orange}{'Select and Finalize Web Series Episode for Analysis'}, then \textcolor{orange}{'Compile Audience Reactions from YouTube Comments'}, and also \textcolor{orange}{'Compile Audience Reactions from Relevant Subreddit'}. \\
Agent1: Task durations are: \textcolor{orange}{Task A} takes \textcolor{blue}{2 hour(s)}, \textcolor{orange}{Task B} takes \textcolor{blue}{1 hour(s)}, \textcolor{orange}{Task C} takes \textcolor{blue}{1 hour(s)}. & Sarah: Okay, so for \textcolor{orange}{'Select and Finalize Episode'}, I'd allocate \textcolor{blue}{2 hours}. Then, \textcolor{orange}{'Compile YouTube Comments'} should take about \textcolor{blue}{1 hour}, and similarly \textcolor{orange}{'Compile Subreddit Comments'} will also be about \textcolor{blue}{1 hour}. \\
Agent2: \textcolor{red}{Task B must begin only after Task A is completed.} \textcolor{red}{Task C must begin only after Task A is completed.} & Michael: And just to confirm, \textcolor{red}{both the YouTube comment compilation and the Subreddit compilation can only start after we've finalized the episode selection}, right? \\
Agent1: My working hours are 11:00–19:00 in AEST (GMT+10). I take a meal break from 15:00 to 16:00. I need a \textcolor{purple}{break of at least 3 hour(s) between any two tasks I perform}. & Sarah: Exactly. My working hours are 11:00 to 19:00 AEST, that's GMT+10. I usually take my lunch around 15:00 to 16:00. Also, I need a \textcolor{purple}{break of at least 3 hours between any two tasks I perform}. \\
Agent2: My working hours are 9:00–17:00 in EST (GMT-5). I take a meal break from 13:00 to 15:00. I need a \textcolor{purple}{break of at least 3 hour(s) between any two tasks I perform}. & Michael: Understood. I'm on EST, GMT-5, working 9:00 to 17:00. My lunch break is typically from 13:00 to 15:00. I also need a \textcolor{purple}{3-hour break between tasks}. \\

\bottomrule
\end{tabular}
}
}
\caption{Line-aligned comparison between the prototype and concretized versions of the same task scheduling dialogue. While both express identical task structures, durations, dependencies, and constraints, the naturalistic surface form in the concretized version led models to reduce its reasoning depth and produce a suboptimal schedule.}
\label{fig:verbatim_comparison_colored}
\end{figure*}

\begin{table*}[htbp]
\centering
\small
\begin{tabular}{p{0.28\textwidth} p{0.32\textwidth} p{0.32\textwidth}}
\toprule
\textbf{Aspect} & \textbf{Prototype Prediction} & \textbf{Concretized Prediction} \\
\midrule
Final predicted completion time & \textbf{2014-12-17 15:00 GMT (correct)} & \textbf{2014-12-18 02:00 GMT (incorrect)} \\
Task scheduling strategy & \textcolor{blue}{Assigned Tasks B and C to different agents to enable parallelism} & \textcolor{red}{Assigned all tasks to one agent, resulting in serial execution} \\
\texttt{thoughts\_token\_count} & \textbf{6145} & \textbf{4673} \\
Constraint enforcement & \textcolor{blue}{All constraints met with minimal delay} & \textcolor{red}{Ignored agent availability and forced unnecessary waiting} \\
Failure point & N/A & Ignored the possibility of parallel task execution \\
Reasoning excerpt & “Agent1 performs Task A at 01:00 GMT. Agent2 starts Task C at 14:00 GMT.” & “Sarah performs Task A at 01:00 GMT. After 3h break and lunch, she performs Task B and then Task C next day.” \\
Reasoning summary & \textcolor{blue}{Used both agents to minimize total time} & \textcolor{red}{Failed to use second agent, increasing project duration} \\
\bottomrule
\end{tabular}
\caption{Despite identical constraints, the model generated correct reasoning in the abstract prototype but failed in the concretized version. The discrepancy corresponds with a significant drop in reasoning token length.}
\label{tab:token_length_impact}
\end{table*}

%% file: error_samples/error_samples_timezone.tex
\paragraph{Time Zone Misinterpretation.}  
We explain this type of error with the example illustrated in Fig.~\ref{fig:timezone_mismatch} and Table~\ref{tab:timezone_mismatch}.
The model incorrectly interpreted \textcolor{purple}{\textbf{Sarah's time zone}} (marked as \texttt{CET}) as UTC+2, likely assuming it referred to daylight saving time (\texttt{CEST}). However, the input explicitly said \texttt{CET}, which is UTC+1. This led the model to assume Sarah could begin work at 08:00 UTC instead of the correct 09:00 UTC. As a result, Task A was scheduled one hour earlier than valid, allowing Tasks 2 and 3 to begin prematurely. This propagation caused the model to report a project finish time of \textbf{13:00 GMT}, which is an hour earlier than the actual earliest feasible time of \textbf{14:00 GMT}. The misalignment is entirely due to the time zone conversion error. See  Fig.~\ref{fig:timezone_mismatch} and Table~\ref{tab:timezone_mismatch} for an illustration. This error highlights a broader challenge: models often rely on heuristics or default assumptions about time zones (which become hallucinations in TCP), making them vulnerable to subtle but consequential temporal misinterpretations even when the input is technically correct.

\paragraph{Project Start Time Violation.}
We explain this type of error with the example illustrated in Fig.~\ref{fig:project_start_violation} and Table~\ref{tab:project_start_violation}.
The model violated the \textbf{project start time} constraint by scheduling Task A at \textcolor{red}{22:00 GMT on October 20}, which is \textit{before} the allowed start of \textbf{00:00 GMT on October 21}. While the time zone conversion from AEST to GMT was correctly handled (Sarah’s 8:00 AM AEST availability does translate to 22:00 GMT), the model failed to enforce the global project start constraint. This early execution allowed subsequent dependent tasks to start and complete earlier than valid. Specifically, it enabled Ben’s Task C to begin at 07:00 GMT and complete by 08:00 GMT. However, once Task A is correctly restricted to start no earlier than 00:00 GMT, all subsequent tasks are pushed back. Accounting for Sarah’s required 3-hour break, Ben’s availability, and lunch hours, the actual earliest valid project completion time becomes \textbf{14:00 GMT on October 21}. The error propagated entirely from violating the global time constraint on project start. See Fig.~\ref{fig:project_start_violation} and Table~\ref{tab:project_start_violation} for illustration. This failure indicates that models can overlook global constraints even when local scheduling appears consistent, suggesting a need for stronger global constraint enforcement mechanisms in temporal planning.

\input{error_samples/time_zone_misinterpretation_table}
\input{error_samples/project_start_time_violation_table}

%% file: error_samples/time_zone_misinterpretation_table.tex
\begin{figure*}[htbp]
\centering
\noindent\fbox{%
\parbox{\textwidth}{%
\small
\textbf{Original Dialogue:}
\begin{quote}
Sarah: Alright Mark, let's get this initiative for the standardized internal project progress report template underway.\\
Mark: Great, Sarah! Could you refresh my memory on how we've structured the initial tasks for this? I know we discussed the breakdown.\\
Sarah: Certainly. We've identified three core tasks to start: first, \textbf{\textcolor{orange}{"Brainstorm Key Data Points for Report"}} to define what information is crucial. Following that, we have \textbf{\textcolor{orange}{"Draft Report Template Layout in Google Docs"}} to create the actual template. And finally, \textbf{\textcolor{orange}{"Develop Submission Guidelines and Review Protocol Document"}} to establish the process around it.\\
Mark: That makes sense. And what are the estimated durations for these specific tasks?\\
Sarah: Good question. According to our planning, \textcolor{blue}{"Brainstorm..."} is set for 1 hour. \textcolor{blue}{"Draft..."} should take about 2 hours, and \textcolor{blue}{"Develop..."} is allocated 3 hours.\\
Mark: Okay, and in terms of dependencies, \textbf{\textcolor{red}{"Draft..."} and "Develop..."} both depend on \textbf{\textcolor{red}{"Brainstorm..."}}.\\
Sarah: Just so you know my schedule, I'm working from \textbf{\textcolor{purple}{10:00 to 18:00 CET}}. I have my standard lunch break between \textbf{\textcolor{purple}{15:00 and 16:00}}. I also need at least a 1 hour break between any two distinct tasks.\\
Mark: Understood. My hours are \textbf{\textcolor{teal}{8:00 to 16:00 UTC}}, and I'll take my lunch from \textbf{\textcolor{teal}{11:00 to 12:00}}. I also require at least 1 hour between my tasks.
\end{quote}
}
}
\caption{Annotated dialogue highlighting time zone (\textcolor{purple}{CET}/\textcolor{teal}{UTC}), task descriptions (\textcolor{orange}{task names}), durations (\textcolor{blue}{durations}), and dependencies (\textcolor{red}{dependencies}).}
\label{fig:timezone_mismatch}
\end{figure*}

\begin{table*}[htbp]
\centering
\small
\begin{tabular}{p{0.28\textwidth} p{0.32\textwidth} p{0.32\textwidth}}
\toprule
\textbf{Aspect} & \textbf{Model's Assumption} & \textbf{Actual Constraint} \\
\midrule
Sarah's timezone & \textcolor{red}{CET = UTC+2} (interpreted as summer time) & \textcolor{blue}{CET = UTC+1} (as stated) \\
Sarah’s working hours (UTC) & 08:00--16:00 & \textbf{09:00--17:00} \\
Start of Task A & 08:00 (Sarah) & \textbf{09:00} \\
Task B and 3 start eligibility & After 09:00 & \textbf{After 10:00} (due to Task A duration = 1h) \\
Earliest Task C window & Sarah: 10:00--13:00 & \textbf{11:00--14:00} (requires 3h block before lunch) \\
\textbf{Project completion time} & \textcolor{red}{\textbf{13:00 UTC}} & \textcolor{blue}{\textbf{14:00 UTC}} \\

\bottomrule
\end{tabular}
\caption{Comparison of model reasoning vs. actual constraint enforcement. Time zone error leads to premature task scheduling.}
\label{tab:timezone_mismatch}
\end{table*}

%% file: error_samples/project_start_time_violation_table.tex
\begin{figure*}[htbp]
\centering
\noindent\fbox{%
\parbox{\textwidth}{%
\small
\textbf{Original Dialogue:}
\begin{quote}
Ben: Alright Sarah, let's nail down the schedule for our `Echoes of Oakwood' project. We need to get this wrapped up quickly.\\
Sarah: Okay, Ben. Could you remind me how we decided to break down the main research tasks for analyzing the Oakwood rezoning coverage?\\
Ben: Sure, Sarah. We agreed on three main stages: First, \textbf{\textcolor{orange}{"Pinpoint and Archive Initial Reports on the Oakwood Rezoning from `The Chronicle' and `Channel 7'"}}. Then, we'll have two parallel analyses: \textbf{\textcolor{orange}{"Deconstruct Narrative Framing \& Source Use in `The Oakwood Chronicle' Rezoning Articles"}} and \textbf{\textcolor{orange}{"Deconstruct Narrative Framing \& Source Use in `Channel 7 Oakwood News' Rezoning Segments"}}.\\
Sarah: Right. \textcolor{blue}{"Pinpoint..."} should take 1 hour. \textcolor{blue}{"Deconstruct...Chronicle"} takes 2 hours, and \textcolor{blue}{"Deconstruct...Channel 7"} takes 1 hour.\\
Sarah: And just to confirm, \textbf{\textcolor{red}{"Deconstruct...Chronicle"}} and \textbf{\textcolor{red}{"Deconstruct...Channel 7"}} both depend on the completion of \textbf{\textcolor{red}{"Pinpoint..."}}.\\
Sarah: My working hours are \textcolor{purple}{8:00 to 16:00 AEST (GMT+10)}. I take lunch between \textcolor{purple}{11:00 and 12:00}, and I need at least a 3-hour break between tasks.\\
Ben: I'm on \textcolor{teal}{UTC time}, working from \textcolor{teal}{7:00 to 15:00}, with lunch from \textcolor{teal}{11:00 to 13:00}. I need at least a 1-hour break between tasks.
\end{quote}
}
}
\caption{Annotated dialogue highlighting time zones (\textcolor{purple}{AEST}/\textcolor{teal}{UTC}), task descriptions (\textcolor{orange}{task names}), durations (\textcolor{blue}{durations}), and dependencies (\textcolor{red}{dependencies}).}
\label{fig:project_start_violation}
\end{figure*}

\begin{table*}[htbp]
\centering
\small
\begin{tabular}{p{0.28\textwidth} p{0.32\textwidth} p{0.32\textwidth}}
\toprule
\textbf{Aspect} & \textbf{Model's Assumption} & \textbf{Actual Constraint} \\
\midrule
Project start time & Tasks may start at \textcolor{red}{2019-10-20 22:00 GMT} & \textcolor{blue}{Project begins no earlier than 2019-10-21 00:00 GMT} \\
Task A scheduled for Sarah & 22:00–23:00 GMT on Oct 20 & \textbf{Must begin \underline{on or after} 00:00 GMT Oct 21} \\
Impact on downstream tasks & Tasks 2 and 3 start earlier due to early Task A & Tasks 2 and 3 must be delayed accordingly \\
Final task end time & \textcolor{red}{08:00 GMT} & \textcolor{blue}{14:00 GMT}\\
\bottomrule
\end{tabular}
\caption{Comparison of model reasoning vs. actual constraint enforcement. Project start time constraint was violated, allowing premature scheduling.}
\label{tab:project_start_violation}
\end{table*}

%% file: reasoning_path_comparison/reasoning_vs_standard.tex
We include illustrations of the qualitative case studies (Fig. \ref{fig:qwen_comparison} and Fig. \ref{fig:gemini_gpt_comparison}; Table \ref{tab:reasoning_mode_impact_qwen} and Table \ref{tab:reasoning_mode_impact_gemini_gpt}) in this section.

\input{reasoning_path_comparison/reasoning_vs_standard_qwen}
\input{reasoning_path_comparison/reasoning_vs_standard_gemini_gpt}

%% file: reasoning_path_comparison/reasoning_vs_standard_qwen.tex
\begin{figure*}[t]
\centering
\noindent\fbox{%
\parbox{\textwidth}{%
\small
\begin{minipage}[t]{0.48\textwidth}
\textbf{Output Trace of Qwen3 14B with Thinking Enabled}\par
\vspace{0.5em}
<think>
Okay, let's try to figure out the earliest completion date for their study. First, I need to outline all the tasks...

\textbf{Alternatively}, assign Task A to Dr. Hanson...\textbf{But} let's try to find the optimal way.
\textbf{If} we assign Task A to Mark Chen...
\textbf{Wait}, but that would be three weekdays, but with a weekend in between...
\textbf{Alternatively}, if he starts Task A on Monday,..
\textbf{But} if Dr. Hanson does Task A, she can work any day,...
</think>

\textbf{1. Task A – Design Priming Materials and Web Interface (3 days):} \\
Assign Dr. Hanson to Task A...

\textbf{2. Task C – Manipulation Check Questions (1 day):} \\
Assign Dr. Hanson to Task C.

\textbf{3. Task B – Recruitment and Data Collection (3 days):}
Assign Mark Chen to Task B.

\textbf{4. Final Completion Date:}\\
Task A is completed on July 20.
Task C is completed on July 20.
Task B is completed on July 23.

\textbf{5. Final Answer:}\\
2014-07-23

\end{minipage}
\hfill
\begin{minipage}[t]{0.48\textwidth}
\textbf{Output Trace of Qwen3 14B with Thinking Disabled}\par
\vspace{0.5em}
\textbf{1. Identify the Tasks and Their Durations:} \\
From the conversation, the project is divided into three main tasks...

\textbf{2. Task Dependencies:} \\
Task B depends on Task A being completed...

\textbf{3. Work Constraints:}\\
Mark Chen:
Works only on weekdays (Monday to Friday)...

\textbf{4. Determine the Earliest Start Date:}\\
The conversation happens on 2014-07-15. That is a Tuesday...

\textbf{5. Assign Tasks to People:}\\
We need to assign the tasks to Mark Chen and Dr. Hanson...

\textbf{Assign Task A to Mark Chen:}\\
Task A is 3 days. Mark Chen can work 3 consecutive weekdays...

\textbf{After Task A is completed (2014-07-22), Tasks B and C can start.}\\
Now, we need to assign Task B (3 days) and Task C (1 day) to Dr. Lena Hanson...

\textbf{6. Final Project Completion Date}\\
... So, the earliest date that the entire project can be completed is: 2014-07-27
\end{minipage}
}
}
\caption{Comparison of output traces generated by Qwen3 14B with thinking enabled and disabled + CoT prompting, for solving the same problem. 
}
\label{fig:qwen_comparison}
\end{figure*}

\begin{table*}[t]
\centering
\small
\begin{tabular}{p{0.28\textwidth} p{0.32\textwidth} p{0.32\textwidth}}
\toprule
\textbf{Aspect} & \textbf{Prototype Prediction} & \textbf{Concretized Prediction} \\
\midrule
Final predicted completion time & \textbf{2014-07-23 (correct)} & \textbf{2014-07-27 (incorrect)} \\
Reasoning strategy & \textcolor{blue}{Enumerated an initial solution, then revises for optimal arrangement} & \textcolor{red}{Assigned tasks without careful analysis and no revision} \\
Failure point & N/A & Assigned task A to Mark Chen on the dates that are weekends. \\
Reasoning excerpt & “Assign Dr. Hanson to Task A.” & “Mark Chen can work 3 consecutive weekdays (e.g., 2014-07-18, 2014-07-21, 2014-07-22) – that’s 3 days.” \\
Reasoning summary & \textcolor{blue}{Activating reasoning mode encourages iterative reflection, which improves the chances of reaching the correct solution} & \textcolor{red}{Failed to convert temporal information with limited reasoning ability} \\
\bottomrule
\end{tabular}
\caption{Comparison of output traces generated by Qwen3 14B with thinking enabled and disabled + CoT prompting, for solving the same problem. }
\label{tab:reasoning_mode_impact_qwen}
\end{table*}

%% file: reasoning_path_comparison/reasoning_vs_standard_gemini_gpt.tex
\begin{figure*}[t]
\centering
\noindent\fbox{%
\parbox{\textwidth}{%
\small
\begin{minipage}[t]{0.48\textwidth}
\textbf{Reasoning Path Structure of Gemini 2.5 flash}\par
\vspace{0.5em}

\textbf{1. Determine the Start Date:} \\
The conversation is on 2014-04-02 (a Wednesday). The project starts "this Saturday," which is 2014-04-05.

\textbf{2. Identify Tasks and Durations:} \\
Task A: Identify Potential Crises \& Audiences (2 days) \\
Task B: Draft Response Protocols \& Templates (1 day) \\
Task C: Establish Channels \& Monitoring Plan (1 day)

\textbf{3. Identify Dependencies:}
Task B depends on Task A. Task C depends on Task B.
The sequence is A -> B -> C.

\textbf{4. Identify Resource Constraints:}\\
\textbf{Chloe:}
Available: Any day (weekdays \& weekends)...\\
\textbf{David:}
Available: Weekdays only (Mon-Fri)...

\textbf{5. Schedule the Tasks}\\
\textbf{Task A (2 days):}\\
Must start on 2014-04-05 (Saturday). David cannot work weekends, so Chloe must do Task A...\\
\textbf{Task B (1 day):}\\
Depends on Task A. Earliest start is Monday, 2014-04-07...\\
\textbf{Task C (1 day):}\\
Depends on Task B. Earliest start is the day after Task B finishes... \\
The final answer is 2014-04-08.

\end{minipage}
\hfill
\begin{minipage}[t]{0.48\textwidth}
\textbf{Reasoning Path Structure of GPT-4.1 mini}\par
\vspace{0.5em}
\textbf{Given:}
Conversation date: 2014-04-02 (Wednesday)\\
Project start: This Saturday (2014-04-05)\\
...

\textbf{1: Assign tasks to Chloe or David}\\
We have 3 tasks: 2 days, 1 day, 1 day... 

\textbf{2: Consider starting date and constraints}\\
Project starts Saturday, 2014-04-05...

\textbf{3: Chloe's sister visit next week (2014-04-07 to 2014-04-13)}\\
Chloe wants a couple of full days off next week (7th to 13th)...

\textbf{4: Plan the schedule}\\
We want to finish as early as possible.

\textbf{5: Assign tasks to minimize total duration}\\
Option 1: Chloe does all tasks. Task 1: 2 days, Task 2: 1 day, Task 3: 1 day, Total: 4 days...

\textbf{6: Can David help to finish earlier?}\\
David can only work weekdays. If David does Task 1 (2 days) on Mon (7th) and Tue (8th), then Chloe can do Task 2 and 3 on Wed (9th) and Thu (10th)...

\textbf{7: Can we finish earlier than Thu (10th)?}
No, because: David cannot work weekends (Sat, Sun)...

\textbf{Final conclusion:}\\
Earliest project completion date is 2014-04-10.
\end{minipage}
}
}
\caption{Comparison of output traces generated by a reasoning model Gemini 2.5 Flash and a standard LLM GPT-4.1 mini prompted with CoT, for solving the same problem. 
}
\label{fig:gemini_gpt_comparison}
\end{figure*}

\begin{table*}[t]
\centering
\small
\begin{tabular}{p{0.28\textwidth} p{0.32\textwidth} p{0.32\textwidth}}
\toprule
\textbf{Aspect} & \textbf{Prototype Prediction} & \textbf{Concretized Prediction} \\
\midrule
Final predicted completion time & \textbf{2014-04-08 (correct)} & \textbf{2014-04-10 (incorrect)} \\
Reasoning strategy & \textcolor{blue}{Solved task-by-task with a structured plan} & \textcolor{red}{Added information incrementally in an ad hoc manner} \\
Failure point & N/A & Unclear subgoal prompts the model to overfit to local context instead of optimizing overall schedule. \\
Reasoning excerpt & “5. Schedule the Tasks: Task A (2 days); Task B (1 day); Task C (1 day)” & “Step 5: Assign tasks to minimize total duration Step 6: Can David help to finish earlier? Step 7: Can we finish earlier than Thu (10th)?” \\
Reasoning summary & \textcolor{blue}{Enabled comprehensive consideration via well-structured reasoning process} & \textcolor{red}{Failed to incorporate critical constraints due to ambiguously named intermediate steps} \\
\bottomrule
\end{tabular}
\caption{Comparison of output traces generated by a reasoning model Gemini 2.5 Flash and a standard LLM GPT-4.1 mini prompted with CoT, for solving the same question. }
\label{tab:reasoning_mode_impact_gemini_gpt}
\end{table*}

%% file: prompt_short.tex
\begin{figure*}[htbp]
\centering
\noindent\fbox{%
\parbox{\textwidth}{%
\small
\textbf{Prompt for \textit{Short} Project Scenario Generation:}
\begin{quote}
You are a project designer. Your job is to invent a new, realistic, and creative project in the area of '\textbf{\{area\}}', inspired by the following example project. Your new project should: \\
- Have a similar number of tasks (with distinct, concrete names and descriptions). \\
- Preserve the same dependency structure between tasks (see below). \\
- Be original, not a copy or trivial rewording of the example or any previously generated project (if provided). \\
- Be plausible for a real team to execute in this domain. \\
- Each task in the project must be scheduled to finish within a FEW HOURS (e.g., 2--6 hours per task, all tasks finish in a single day). \\
\\
Here are the previously generated projects. The new project MUST NOT copy or trivially reword any of them. Use them as negative examples to ensure originality:\\
\textbf{\{prev\_refs\}}\\

Example project (format):\\
\textbf{\{example\_json\}}\\
\\
---

Please output a new project as a JSON object in the EXACT SAME FORMAT as the example above, with all the following fields: \\
- project\_name: (str) The new project name \\
- tasks: (list of dict) Each with 'task\_name' and 'possible\_contents' \\
- dependencies: (list of [from\_task, to\_task]) \\
- task\_name\_map: (dict) Mapping new task names to generic names (e.g., 'Task A', 'Task B', ...) \\
- dependencies\_map: (list of [from\_task\_generic\_name, to\_task\_generic\_name]), mapping each new task name to its corresponding generic task name (e.g., 'Task A', 'Task B', ...) from dependencies \\

IMPORTANT: Only output the JSON object, no commentary or markdown. Maintain the same field names and structure as the example. \\
\end{quote}
}
}
\caption{Complete prompt for project scenario generation for \textit{short} problems. \textbf{\{area\}} denotes a domain, e.g., Computer Science. \textbf{\{prev\_refs\}} represents the generated project scenarios within the same domain and dependency type from rounds $0$ to $r-2$. \textbf{\{example\_json\}} is the project scenario within the same domain and dependency type generated from the previous round $r-1$.}
\label{fig:prompt_scenario_short}
\end{figure*}